\documentclass[11pt]{article}

\usepackage[final]{acl}

\usepackage{times}
\usepackage{latexsym}

\usepackage[T1]{fontenc}
\usepackage[utf8]{inputenc}

\usepackage{microtype}
\usepackage{xspace}
\usepackage{amsmath, amssymb}
\usepackage{graphicx}

\usepackage{inconsolata}

\usepackage{graphicx}

\usepackage[utf8]{inputenc} % allow utf-8 input
\usepackage[T1]{fontenc}    % use 8-bit T1 fonts
\usepackage{hyperref}       % hyperlinks
\usepackage{url}            % simple URL typesetting
\usepackage{booktabs}       % professional-quality tables
\usepackage{amsfonts}       % blackboard math symbols
\usepackage{nicefrac}       % compact symbols for 1/2, etc.
\usepackage{microtype}      % microtypography
\usepackage{xcolor}         % colors
\usepackage[table]{xcolor}
\usepackage[dvipsnames]{xcolor}
\usepackage[most]{tcolorbox}
\usepackage{graphicx}
\usepackage{wrapfig}
\usepackage{amsmath}
\usepackage{natbib}
\usepackage{xspace}
\usepackage{amssymb}
\usepackage{float}
\usepackage[normalem]{ulem}
\usepackage{algorithm}
\usepackage{cleveref}
\usepackage{threeparttable}
\usepackage[normalem]{ulem}
\usepackage[most]{tcolorbox}
\usepackage{multirow}
\usepackage{makecell}
\usepackage{bm} 
\usepackage{enumitem}
\usepackage[most]{tcolorbox}
\tcbuselibrary{listingsutf8}
\usepackage{longtable}
\usepackage{array}
\usepackage{algpseudocode}
\usepackage{amsthm}
\usepackage{cleveref}
\usepackage{lipsum}
\definecolor{lightbluepurple}{RGB}{235, 240, 250}
\definecolor{MyHighlight}{RGB}{245, 169, 169}

\newcommand{\fakeparagraph}[1]
{\vspace{0.5mm}\noindent\textbf{#1}}
\newcommand{\eg}{e.g.,}

\newcommand\figref[1]{Figure~\ref{#1}}

\newcommand\tabref[1]{Table~\ref{#1}}

\newcommand\secref[1]{\S\ref{#1}}

\newcommand{\figrefsub}[2]{\hyperref[#1]{\ref{#1}\hspace{1.5pt}(#2)}}
\author{Chengzhengxu Li\textsuperscript{$1$}, Xiaoming Liu\textsuperscript{$1,\ast$}, Zhaohan Zhang\textsuperscript{$2$}, Shengchao Liu\textsuperscript{$1$},\\ \textbf{Guoxin Ma\textsuperscript{$1$}, Yu Lan\textsuperscript{$1$}, Cong Wang\textsuperscript{$3$},  Chao Shen\textsuperscript{$1$}} \\
        \textsuperscript{1}Faculty of Electronic and Information Engineering, Xi'an Jiaotong University\\ 
        \textsuperscript{2}Queen Mary University of London, \textsuperscript{3}City University of Hong Kong\\
        \textsuperscript{$\ast$} Corresponding author\\
        \texttt{ 
        \{czx.li, liusc, guoxin.ma\}@stu.xjtu.edu.cn}, zhaohan.zhang@qmul.ac.uk\\
        \texttt{
        \{xm.liu, chaoshen, ylan2020\}@xjtu.edu.cn}, congwang@cityu.edu.hk\\
        }

\title{Can Reasoning Path still be Effective as Input? Bridging Post-Reasoning to Chain-of-Thought Compression}
\begin{document}
\maketitle
\begin{abstract}\label{abstract}
Recent developments have enabled advanced reasoning in Large Language Models (LLMs) via long Chain-of-Thought (CoT), trading efficiency during inference for performance.
Existing works focus on compressing generated CoT in reasoning, which impairs the necessary information for deriving the correct answer.
In this work, we propose post-reasoning, a reasoning paradigm that takes CoT as a part of context to simplify the reasoning task for LLMs.
We find that post-reasoning significantly reduces the generation length of LLMs, but its effectiveness hinges on the efficiency and the reliability of the contextual CoT generation.
% while long CoT suffers from high computational costs and significant latency losses owing to the autoregressive nature of generative LLMs.
% CoT compression aims to improve efficiency in the reasoning process by reducing output length.
% Previous works trade reasoning efficiency by either laborious discrete prompt designing or the construction of external compressed CoT datasets that sacrifice key reasoning details.
Therefore, we propose Upfront CoT (UCoT), an efficient post-reasoning framework for CoT compression. 
UCoT trains a lightweight model (compressor) to provide contextual CoT in form of soft tokens and trains the LLM (executor) to leverage this contextual CoT for producing the final answer.
Extensive experiments show that UCoT maintains the powerful reasoning ability of executor while significantly reducing the length of CoT. 
It is worth mentioning that when applying UCoT to the Qwen2.5-7B-Instruct model, the usage of tokens on GSM8K dataset is reduced by 50\%, while the performance is 3.08\% higher than that of the state-of-the-art (SOTA) method. 
%The code is available at: \url{https://github.com/czx-li/UCoT}.
\end{abstract}

\section{Introduction}
Chain-of-Thought (CoT) reasoning has emerged as a pivotal paradigm for overcoming the bottleneck of complex reasoning capabilities in large language models (LLMs) \cite{wei2022chain,sprague2025cot}.
Recent advancements exemplified by models such as DeepSeek-R1 \cite{guo2025deepseek} and OpenAI's o3 \cite{el2025competitive}, demonstrate that scaling test-time computation on reasoning is a key path to advanced intelligence \cite{snell2024scaling}.
This technique strengthens the model’s ability to reason from scratch by decomposing complex problems into sequential steps, but it brings substantial computational overhead due to the autoregressive nature of current LLMs \cite{su2025between, arora2025training}.
Existing works about compressing the CoTs focus on simplifying model output through prompting strategies \cite{xu2025chain, aytes2025sketch} or post-training \cite{kang2025c3ot, xia2025tokenskip} without providing any complementary contextual information, which leads to efficiency gains at the cost of significant performance degradation \cite{zhang-etal-2025-lightthinker}.

% As illustrated in \textcolor{red}{Figure}, the inference latency is dominated by token generation rather than context processing because the generation is a strictly sequential decoding process, whereas processing context can be handled in parallel during the prefill stage \cite{zhang2022automatic}.
\begin{figure}[!t]
    \centering
    \resizebox{0.45\textwidth}{!}{%
        \includegraphics{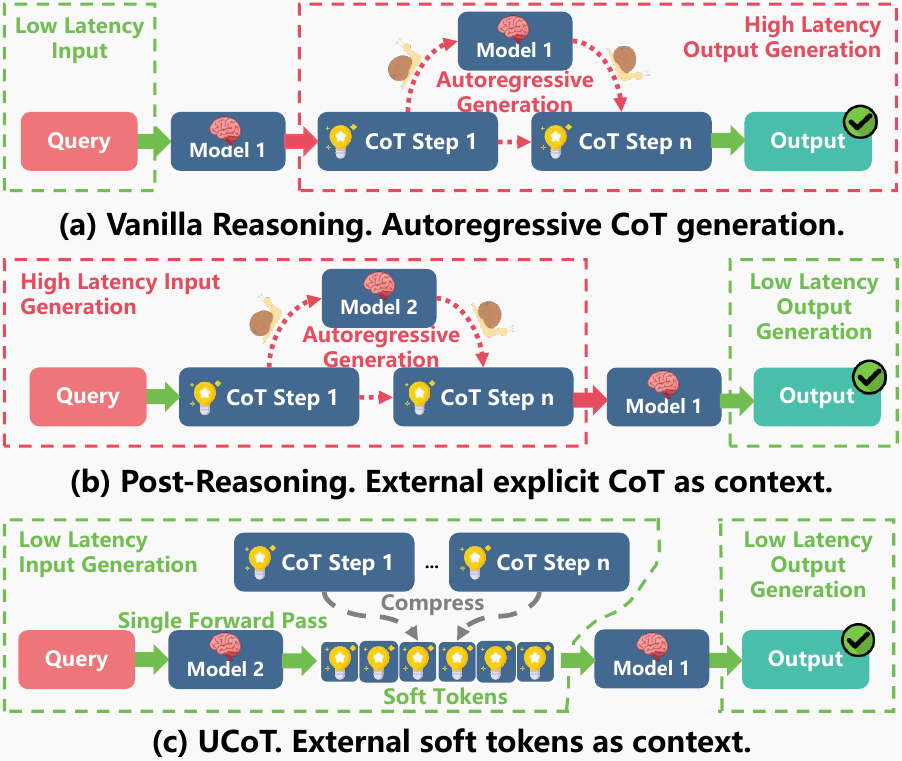} 
    }
	\caption{Comparison of reasoning paradigms. (a) Vanilla Reasoning generates CoT autoregressively, causing high output latency. (b) Post-Reasoning uses explicit CoT as input, trading contextual CoT generation for short output. (c) UCoT generates soft tokens via a single forward pass, achieving low latency and high performance.}
    \label{fig:intro}
\end{figure}
%In this work, we revisit the role of CoT as the premise for deriving the final answer and introduce the concept of {\color[RGB]{0, 81, 186}{\textbf{\textit{post-reasoning}}}}, where an LLM continues its reasoning based on contextual CoT (i.e., CoT provided as part of the model input).
%As illustrated in Figure \ref{fig:intro} (b), 
%CoT conveys the same information needed for the final answer, regardless of whether it is generated during decoding or supplied as context.
%Moreover, the inference latency in autoregressive LLMs is dominated by token generation rather than context processing because the generation is a strictly sequential decoding process, whereas processing context can be handled in parallel during the prefill stage \cite{zhang2022automatic}.
%Motivated by this observation, we ask: \textit{“Can post-reasoning alleviate the need for LLMs to overthink and reduce inference latency?”}
%We further validate the feasibility of this paradigm through a pilot study (Section 3) and pose two finding regarding post-reasoning scheme:
In this work, we revisit the role of CoT as the premise for deriving the final answer.
As illustrated in Figure \figrefsub{fig:intro}{a}, we point that CoT is not merely a sequence of outputs, but effectively a self-generated input context that the LLM produces to bridge the logical gap between the input query and the final answer \cite{zhang2022automatic, kojima2022large, wurethinking}.
Motivated by this observation, we first investigate a critical research question to CoT reasoning: \textit{“Can providing reasoning information as input reduce CoT generation in LLMs?"}
Toward this end, we introduce the paradigm of {\color[RGB]{0, 81, 186}{\textbf{post-reasoning}}} as illustrated in Figure \figrefsub{fig:intro}{b}, where an LLM continues its reasoning based on contextual CoT (i.e., CoT generated by a lightweight external model and provided as part of the LLM input).
We further validate the feasibility of this paradigm through a pilot study (Section~\ref{sec:pilot_study}) and pose two critical factors influencing the performance of post-reasoning:
\begin{tcolorbox}[
    colback=gray!5, 
    colframe=gray!20, 
    arc=3pt, 
    left=5pt, right=5pt, top=2pt, bottom=2pt
]
\begin{description}[
    leftmargin=2.0em, % 控制第二行文字缩进的位置，使其对齐第一行正文
    labelsep=0.3em,   % 缩短 F1 到文字之间的距离
    nosep,            % 让行与行之间非常紧凑
    font=\normalfont  % 保持标签字体与正文一致
]
    \item[\boldmath$\mathcal{F}_1$:] Post-reasoning trades contextual CoT generation for LLM shortened output.
    \item[\boldmath$\mathcal{F}_2$:] Post-reasoning performance hinges on the quality of contextual CoTs.
\end{description}
\end{tcolorbox}

The dependency of post-reasoning on efficiently generating reliable contextual CoT motivates us to propose
% concise and reliable contextual CoT in post-reasoning raises the key challenge: how to generate high-fidelity, task-aligned CoTs with minimal overhead and negligible performance loss.
% While post-reasoning reduces inference cost by reusing externally provided CoT, its efficacy critically depends on the quality and efficiency of the injected reasoning context raising the key challenge: how to generate high-fidelity, task-aligned CoTs with minimal overhead and negligible performance loss.
\textbf{U}pfront \textbf{C}hain-\textbf{o}f-\textbf{T}hought (UCoT) as shown in Figure \figrefsub{fig:intro}{c}, a post-reasoning framework designed to achieve an optimal trade-off between efficiency and effectiveness in reasoning.
UCoT follows the post-reasoning paradigm, alleviating the need to generate lengthy reasoning paths by supplying compressed contextual CoT.
Specifically, UCoT \textbf{constructs contextual CoT as a sequence of soft tokens} produced by a lightweight model (compressor), and performs post-reasoning using a large model (executor).
We first distill the reasoning capability of the large model into the compressor, and train the compressor to encode reasoning processes into soft tokens.
The executor is then trained to leverage this compressed contextual CoT to generate concise yet accurate responses.
Experimental results demonstrate that UCoT compresses CoT outputs while preserving LLM reasoning ability, and outperforms baselines and SOTA on multiple tasks. For instance, UCoT helps Qwen2.5-7B-Instruct cut 50\% of required tokens with only a 5.62\% performance drop on GSM8K, obtaining 3.08\% better than SOTA method Tokenskip \cite{xia2025tokenskip}.
Our contributions are as follows: 
\begin{itemize}
    \item \textbf{Post-Reasoning.} 
    We propose post-reasoning as a novel reasoning paradigm where the model continues reasoning conditioned to contextual CoT. 
    It reveals the feasibility of breaking the latency bottleneck in autoregressive models by contextualizing the reasoning path.

    \item \textbf{Efficient Reasoning Framework.}
    We introduce UCoT, a post-reasoning framework that takes a sequence of soft tokens as contextual CoT and reducing the generation length of LLMs. 
    Through two-stage training, UCoT enables a lightweight compressor to inject high-quality contextual CoT that guides efficient and accurate reasoning.
    \item \textbf{Outstanding Performance.} Our method significantly outperforms baseline and SOTA methods on two open-source models and two public datasets. Furthermore, we also demonstrate its excellent generalization ability to currently popular large reasoning models and more challenging datasets.
\end{itemize}

\section{Related Works}
\noindent \textbf{CoT Reasoning in LLMs.}
With the broader application of LLMs in logistically-complex domain such as 
mathematical study \cite{setlur2024rl}, code generation \cite{fakhoury2024llm}, and scientific question answering \cite{zhuang2023toolqa}, it becomes essential to enhance the reasoning ability of LLMs \cite{liu2024exploring, ramprasad2024analyzing}.
Researchers have proposed the Chain-of-Thought (CoT) reasoning method \cite{wei2022chain,wang2022self}, which enhances model performance by explicitly generating intermediate reasoning steps through LLMs, to unlock the potential of LLMs.
Subsequent studies built upon CoT by employing multi-step sampling \cite{yu2023thought,yao2023tree} or multi-path summarization \cite{chen2024boosting,sun2024enhancing} to aggregate information from multiple reasoning trajectories, thereby further enhancing LLMs' reasoning capabilities in complex problem-solving.
%Furthermore, recently released reasoning-focused language models, such as DeepSeek-R1 \cite{guo2025deepseek}, Qwen2.5 \cite{hui2024qwen2} and OpenAI's o3 \cite{el2025competitive}, integrate CoT with language model training using algorithms like GRPO \cite{shao2024deepseekmath}, demonstrating remarkable advancements in handling complex tasks. 
However, since transformer-based LLMs generate text autoregressively, their processing time for complex problems increases linearly with the length of the CoT \cite{nayab2024concise}. Therefore, our work focus on reducing the length of CoT generated by LLMs while maintaining their reasoning performance.

\noindent \textbf{CoT Compression.}
% While CoT reasoning effectively enhances language models' capability in handling complex tasks, its lengthy reasoning process significantly increases latency and computational costs. 
% To address this challenge, 
CoT compression aims to reduce the output length of the modern reasoning models while preserving reasoning accuracy.
%, thereby enabling high-performance inference in LLMs.
Numerous studies employ carefully designed discrete prompts to guide LLMs toward generating more concise reasoning paths at the output stage \cite{xu2025chain,han2024token}.
However, due to the discrete nature of text, the discrete prompts cannot be directly optimized with gradient descent, necessitating extensive manual design \cite{aytes2025sketch} or complex gradient-free optimization methods \cite{li2025syzygy}.
Another line of research focuses on fine-tuning LLMs on datasets with short but high-quality CoT reasoning paths \cite{cui2025stepwise, xia2025tokenskip}.
However, the cost-effective construction of high-quality concise CoT datasets remains an outstanding challenge.
% For instance, \cite{cui2025stepwise} employs the LLM's perplexity metric to identify and extract pivotal reasoning steps from lengthy CoT sequences for dataset construction. 
% And \cite{xia2025tokenskip} employ the LLMLingua-2 method \cite{pan2024llmlingua} with variable compression ratios. 
% Although these methods eliminate the need for additional modules in LLMs during inference, the cost-effective construction of high-quality concise CoT datasets remains an outstanding challenge.
%In addition, \citet{cheng2024compressed,saunshi2025reasoning} compress each CoT step into continuous latent representations %without \textcolor{red}{the participation of discrete CoT}
%, thereby reducing explicit generation by LLMs at the cost on performance degradation. 
In addition, \citet{cheng2024compressed,saunshi2025reasoning,he2025semcot,zhang-etal-2025-lightthinker} compress each CoT step into continuous latent representations, thereby limiting explicit generation by LLMs at the output stage while incurring performance degradation.
% However, these methods predominantly rely on continuous latent representations to provide cues for LLMs, lacking effective interaction between continuous and discrete reasoning, leading to substantial degradation in LLMs' reasoning performance.
%In contrast, our work focuses on employing continuous prompts to assist LLMs in generating more concise CoT explicitly with the small and large model cooperation, thereby discarding Information loss caused by the construction of external compressed CoT datasets and mitigating performance drop.
In contrast, our work shifts the focus to the LLM input, using contextual CoT to help LLMs shorten reasoning before generation. This enables the spontaneous production of concise CoTs, avoiding information loss from external dataset construction or limiting explicit generation.
In Appendix \ref{B}, we also provide the formal definitions of Chain-of-Thought Compression.

%\noindent \textbf{Small Model Empowerment for LLMs.}
%Enhancing LLMs with small models is trending as an efficient alternative to fine-tuning large models on the downstream tasks \cite{schick2023toolformer,xie2023doremi}. 
%\citet{shen2023hugginggpt} employs collaborative interaction between large and small models to handle complex tool scheduling tasks. 
%Furthermore,  \citet{wang2023tabi} establishes a multi-level reasoning algorithm through language models of varying scales to optimize the inference performance of a single LLM.
%The latest concurrent work \cite{liu2025thoughtmanipulationexternalthought} verifies the possibility of efficient reasoning in LLMs by utilizing CoT generated by small models.

%Building upon this foundation, we extend this paradigm to the domain of CoT compression by employing small models to rapidly generate the continuous representation with rich reasoning information and teaching LLMs to use it.

\section{Preliminary}\label{sec:pilot_study}
In this section, we formalize the post-reasoning paradigm and conduct a pilot study to analyze the impact of post-reasoning on the output length and performance of LLM.
% From our insight, we analyze the impact of contextual CoT during inference across two dimensions: (i) their ability to curtail the autoregressive token count, and (ii) their influence on the model's final reasoning performance.
\subsection{Paradigm Formulation}
\fakeparagraph{Vanilla Reasoning.}
Given an input query $Q= \{ q_{1}, q_{2}, ...,q_{\left | Q \right |}  \} $ consisting of $\left | Q \right |$ tokens, vanilla reasoning process of the LLM is formulated as:
\begin{equation}\label{1}
\{C, A\} = \text{LLM}(Q),
\end{equation}
where $C= \{ c_{1}, c_{2}, ...,c_{\left | C \right |}  \}$ is the CoT with length $|C|$ and
$A= \{ a_{1}, a_{2}, ...,a_{\left | A \right |}  \}$ is the final answer with length $|A|$. 
The CoT $C$ is generated first, followed by the final answer $A$.
Since $|C| \gg |A|$ in most cases, inference latency is primarily dominated by the generation of the CoT rather than the final answer \cite{pope2023efficiently}.

\fakeparagraph{Post-Reasoning.}\label{PR}
In post-reasoning paradigm, the contextual CoT $C'$ generated by an external LLM is treated as a part of the input, which is formulated as:
\begin{equation}\label{2}
\{\hat{C},\hat{A}\} = \text{LLM}(Q \oplus C'),
\end{equation}
where $\hat{C}$ and $\hat{A}$ represents the model's output CoT
and answer conditioned on the contextual CoT $C'$, respectively. 
The $\oplus$ denote the concatenation of query $Q$ and contextual CoT $C'$. 
We compare the length of $|\hat{C}| + |\hat{A}|$ with $|C| + |A|$ generated by Vanilla Reasoning to measure if the model's output can be reduced when given the contextual CoT $C'$. 
Based on this setup, we subsequently explore how post-reasoning affects both output lengths and final answer accuracy in the pilot study.
%We note that post-reasoning is more efficient than traditional reasoning when $|C| \gg |A|$ holds, as inference latency in autoregressive models scales linearly with the length of the generated output rather than the input \textcolor{red}{citation}.
%We subsequently explore how post-reasoning affects the output lengths and final answer in the pilot study.

\begin{figure}[!t]
    \centering
    \resizebox{0.48\textwidth}{!}{%
        \includegraphics{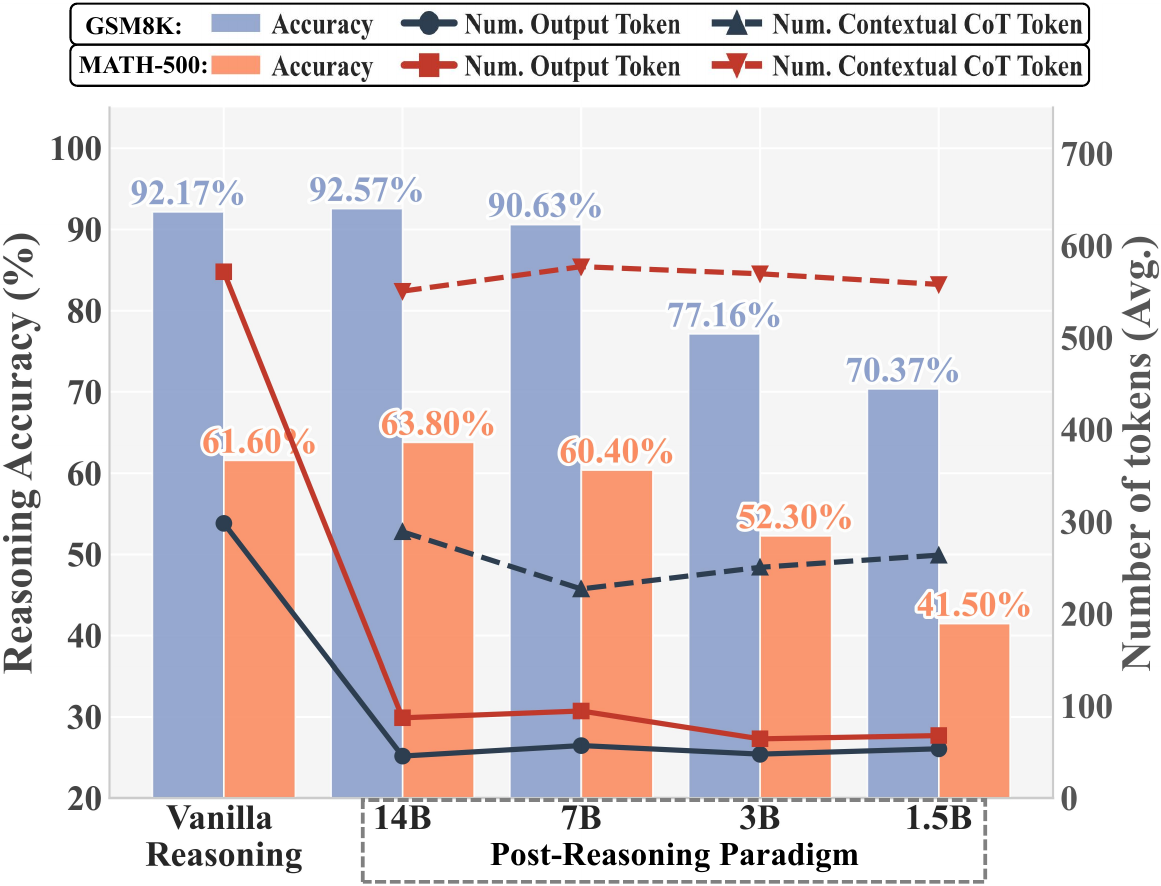} 
    }
	\caption{The impact Post-reasoning on the output length and performance. Post-reasoning facilitates reasoning tasks by trading upfront contextual CoT generation for a significant reduction in output tokens.\protect\footnotemark.}
    \label{fig:pilot}
\end{figure}
\footnotetext{We employ Qwen2.5-{1.5B, 3B, 7B, 14B}-Instruct for constructing contextual CoT and test on the GSM8K \cite{2021arXiv211014168C} and Math-500 \cite{hendrycks2measuring} datasets. }

\subsection{How Post-Reasoning Impacts Reasoning?}
We conduct experiments with Qwen2.5-7B-Instruct \cite{hui2024qwen2}. 
The model takes the query and CoT from external models of different scales as context to derive the output.
We measure the average output length and calculate the answer accuracy for evaluating the paradigm performance.
\begin{figure*}[ht]
    \centering
    \resizebox{0.9\textwidth}{!}{%
        \includegraphics{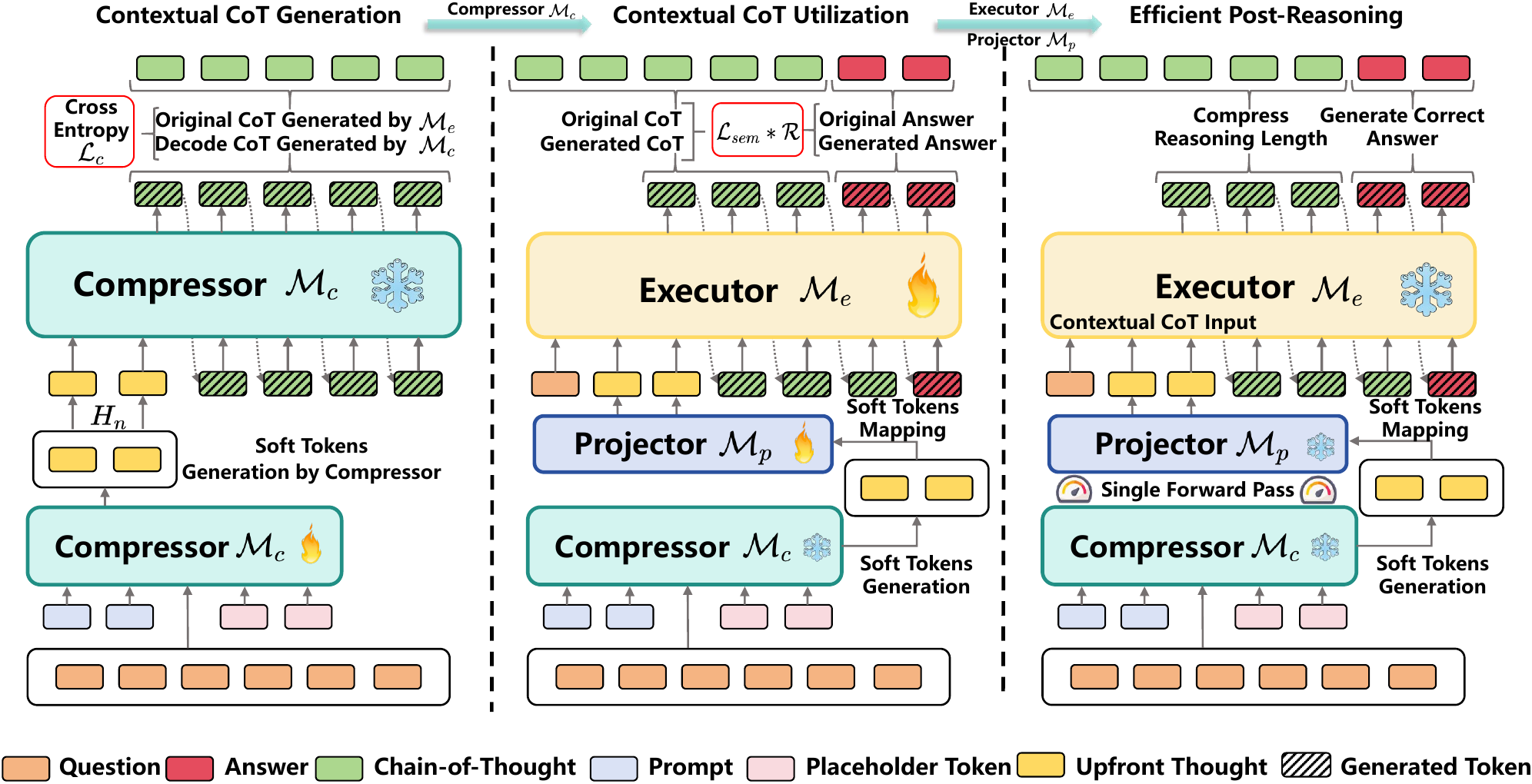} 
    }
    \caption{Overview of Upfront Chain-of-Thought. In contextual CoT generation stage (\secref{section:3.1}) compressor is trained $\mathcal{M}_c$ to generate soft tokens. In contextual CoT utilization stage (\secref{section:3.2}), the executor $\mathcal{M}_{e}$ learns to give correct responses with less CoT length based on the contextual soft tokens. In Efficient Post-Reasoning satge (\secref{section:3.3}), the compressor $\mathcal{M}_c$ and the executor $\mathcal{M}_{e}$ cooperate under the post-reasoning paradigm to achieve efficient reasoning.}
    \label{fig:2} 
\end{figure*}

As shown in Figure~\ref{fig:pilot}, our pilot study unveils following insights:
\textit{i) The post-reasoning paradigm significantly reduces the LLM’s output length but allocates more computation to contextual CoT generation.}
Across both datasets, output tokens are reduced by over 80.00\% relative to vanilla reasoning.
However, this gain relies on generating contextual CoT of comparable scale.
% This suggests that providing additional information as context helps reduce generation.
\textit{ii) Post-Reasoning performance varies with the quality of contextual CoT.}
When CoTs generated by smaller models (e.g., Qwen2.5-1.5B) are used as context, model accuracy drops sharply by approximately 20.00\%, highlighting the necessity of reliable and informative contextual CoT.
These findings validate the feasibility of post-reasoning as an efficient reasoning paradigm, and further reveal that the length and quality of the contextual CoT are two critical factors governing its efficiency and performance.

\section{Efficient Reasoning with Upfront Chain-of-Thought}\label{section:3}
Based on both insights, we propose the Upfront Chain-of-Thought (UCoT) framework to eliminate the autoregressive bottleneck of post-reasoning for contextual CoT.
By transforming explicit reasoning paths into soft tokens via a single forward pass, UCoT achieves significant acceleration while maintaining high-fidelity logic. 
As illustrated in Figure~\ref{fig:2}, our pipeline facilitates a cooperative workflow between a lightweight \textbf{compressor} (for generating soft tokens) and a powerful \textbf{executor} (for deriving the final answer).

\subsection{Contextual CoT Generation}\label{section:3.1}
The goal of this stage is to train a lightweight compressor $\mathcal{M}_{c}$ to efficiently \textbf{condense complex contextual CoTs into a compact sequence of soft tokens}.
Since lightweight models naturally lag behind larger models in reasoning capabilities \cite{wu2024inference}, we distill the reasoning ability of larger model into the compressor.

% a knowledge distillation strategy. 
% We treat the larger executor $\mathcal{M}_{e}$ as the teacher to guide the compressor in encoding high-quality reasoning patterns into the latent space.

\fakeparagraph{Data Construction.}
We derive high-quality training data from the executor itself to ensure the reasoning style aligns with the downstream inference model. 
Given a dataset of query-answer pairs $\{(Q_{n}, A_{n})\}_{n=0}^{N-1}$, we employ the executor $\mathcal{M}_{e}$ to generate the ground-truth CoT $C_n$ for each question $Q_n$:
\begin{equation}\label{eq:teacher_gen} 
C_{n} = \mathcal{M}_{e}(Z_{e} \oplus Q_{n}), 
\end{equation}
where $Z_{e}$ is the system prompt for the executor, and $\oplus$ denotes concatenation. This yields a training corpus $\mathcal{D}=\{(Q_{n}, C_{n}, A_{n})\}_{n=0}^{N-1}$.

\fakeparagraph{Contextual CoT Compression.}
To avoid the latency of autoregressive decoding, we train the compressor to encode the reasoning path into compact soft tokens.
We append a sequence of placeholder tokens $S_{spec}=\{[\text{ucot}]_{m}\}_{m=0}^{M-1}$ of length $M$ to the input.
Instead of generating text, the compressor $\mathcal{M}_{c}$ processes the input in a single forward pass, and we extract the last hidden states corresponding to these placeholders as the soft tokens:
\vspace{-0.1cm}
\begin{equation}\label{eq:input_form} 
X_{c}^{n} = (Z_{c} \oplus Q_{n} \oplus S_{spec}), 
\end{equation} 
\begin{equation}\label{eq:ut_extract} 
H_{n} = \mathcal{M}_{c}(X_{c}^{n}; \theta_{c})=\{h_{m}\}_{m=0}^{M-1}, 
\end{equation}
where $\theta_{c}$ represents the learnable adaptor parameters (e.g., LoRA \cite{hu2022lora}) injected into the compressor, and $H_{n}$ is the resulting soft tokens.
Formally, $H_n$ functions as the latent equivalent of $C'$ in Eq. \ref{2}, allowing the subsequent executor to ingest the reasoning context in a compressed format.

\fakeparagraph{Compressor Optimization.}
To ensure $H_{n}$ captures the semantic essence of the reasoning path, we train the adaptor $\theta_{c}$ to reconstruct the original high-quality CoT $C_n$ from these embeddings.
Following the prompt tuning paradigm \cite{lester2021power,kuratov2025cramming}, we maximize the likelihood of the target CoT conditioned on the soft tokens:
\begin{equation}\label{eq:loss_c}
\mathcal{L}_{c} = \mathbb{E}_{\mathcal{D}}(- \log\mathrm{P}_{\mathcal{M}_{c}}(C_n|H_{n})).
\end{equation}
By minimizing $\mathcal{L}_{c}$, we force the compressor to squeeze the extensive reasoning information of $C_n$ into the dense representation $H_{n}$.
We present the pseudocode for training the compressor $\mathcal{M}_{c}$ in Appendix \ref{A.6}.

\subsection{Contextual CoT Utilization}\label{section:3.2}
The soft tokens $H_n$ need to be aligned with the embedding space of the executor to enable effective utilization by $\mathcal{M}_{e}$ .
Therefore, we use a projector $\mathcal{M}_{p}$, a two-layer MLP that maps $H_n$ into $\mathcal{M}_{e}$'s embedding space.
However, simple alignment is insufficient for effective compression. 
As noted in prior work \cite{xu2025softcot}, LLMs before training often fail to directly leverage such compressed priors to actually reduce generation length or guide accurate reasoning.
To address this, we introduce a reward-guided utilization mechanism that compels the executor to substitute explicit tokens with the provided soft token priors.
Specifically, we constrain the reasoning process by imposing a limited decoding budget on the executor's output.
This constraint yields a truncated residual CoT, denoted as $\bar{C}_{n}$, and is formulated as:
\vspace{-0.4em}
\begin{align}\label{10} 
\bar{C}_{n} = \text{Cutoff}(\mathcal{M}_{e}(\mathcal{M}_{p}(H_{n})\oplus Q_{n};\theta_{e}))_{\alpha},
\end{align}
where $\text{Cutoff}(\cdot)_{\alpha}$ truncates the explicit CoT to a compression ratio $\alpha$, compelling the model to rely on $H_n$ for the missing logic.

\fakeparagraph{Executor Optimization.} 
The executor is optimized towards ensuring both decoding complete semantic information from the contextual CoT and correctly answering the question.
First, we design the semantic loss $\mathcal{L}_{\text{sem}}$ to align the representation of the whole reasoning path, i.e., soft tokens and the executor's generation $\bar{C}_{n}$ before answer with the original long CoT $C_n$ in the executor's final hidden state:
\begin{equation}\label{11}
\begin{gathered}
    \mathcal{L}_{\text{sem}} =\mathbb{E}_{D}(\text{Dist}(H_{\text{UCoT}}, H_{\text{CoT}})) , \\
    H_{\text{UCoT}} = f_{\mathcal{M}_{e}}(\mathcal{M}_{p}(H_{n})\oplus\bar{C}_n), \\
    H_{\text{CoT}} = f_{\mathcal{M}_{e}}(C_n),
\end{gathered}
\end{equation}
where $f_{\mathcal{M}_{e}}(\cdot)$ extracts the last hidden state and $\text{Dist}(\cdot)$ is the standard Mean Absolute Error function.
Second, to ensure the soft tokens guide the model toward the correct answer, we introduce a reward factor $\mathcal{R}$.
This factor penalizes discrepancies in answer confidence between the compressed and original inference:
\begin{equation}\label{12}
\begin{gathered}
    \mathcal{R} =\mathbb{E}_{D}((r_{\text{UCoT}}-r_{\text{CoT}})^2),\\
    r_{\text{UCoT}} = -\log\mathrm{P}_{\mathcal{M}_{e}}(A_n|\mathcal{M}_{p}(H_{n}); \bar{C}_n; \theta_{e}), \\
    r_{\text{CoT}} = -\log\mathrm{P}_{\mathcal{M}_{e}}(A_n|C_n),
\end{gathered}
\end{equation}
The final objective $\mathcal{L}_{e} = \mathcal{L}_{\text{sem}} \cdot \mathcal{R}$ ensures that the executor effectively decodes the soft tokens to recover the full reasoning semantics while maintaining high confidence in the correct answer.
The training pseudocode is provided in Appendix \ref{A.6}.

\subsection{Efficient Post-Reasoning}
\label{section:3.3}
During inference, UCoT operates under the post-reasoning paradigm via a two-stage collaborative pipeline.
First, the compressor $\mathcal{M}_c$ distills the reasoning logic into a sequence of soft tokens.
Next, the executor $\mathcal{M}_e$ ingests these tokens  as contextual CoT $C'$ in Section~\ref{PR} combined with the query to efficiently derive the answer.
Formally, given a query $Q_{n}$, the inference process is formulated as:
\vspace{-0.4em}
\begin{equation}\label{14}
\begin{gathered}
    C'_n =\mathcal{M}_{p}(H_{n})= \mathcal{M}_{p}(M_{c}(Z_c\oplus Q_{n}\oplus S_{spec})), \\
    \{\hat{C}_{n},\hat{A}_{n}\} = \mathcal{M}_{e}(Q_{n}\oplus C'_n),
\end{gathered}
\end{equation}
where the executor's output consists of a concise reasoning path $\hat{C}_{n}$ and the final answer $\hat{A}_{n}$.

\section{Experiments}\label{Experiments}
To verify UCoT's effectiveness, we conduct extensive experiments using Llama-3.1-8B-Instruct \cite{2024arXiv240721783G} and Qwen2.5-7B-Instruct \cite{hui2024qwen2} as executors, and Qwen2.5-1.5B-Instruct \cite{hui2024qwen2} as the compressor with two mathematical reasoning benchmark datasets: GSM8K \cite{2021arXiv211014168C} and Math-500 \cite{2021arXiv210303874H}.
For comprehensive details on the datasets, the implementation of our method, and the training and evaluation protocol, please refer to Appendix~\ref{A.1}, \ref{A.4}, and \ref{A.5}.
% For a detailed description of the datasets and implementation for our method, please refer to Appendix \ref{A.1} and \ref{A.4}.

%%%%%%%%%%%%% Table %%%%%%%%%%%%%
\begin{table*}[!h]
\centering
\renewcommand\arraystretch{0.7}
\resizebox{1\textwidth}{!}{
\begin{tabular}{cc|lccc|lccc}
\toprule \midrule
        &  \multicolumn{1}{c}{}  &  \multicolumn{4}{c}{\textbf{GSM8K}}   & \multicolumn{4}{c}{\textbf{MATH-500}} \\ \cmidrule(lr){3-6} \cmidrule(lr){7-10}
\textbf{Methods} & \multicolumn{1}{c}{\textbf{Ratio}}       & Acc.\ (\%)\ $\uparrow$ & \ Tokens\ $\downarrow$ & \multicolumn{1}{c}{ Latency\ (s)\ $\downarrow$ }& ActRatio\ $\downarrow$ & Acc.\ (\%)\ $\uparrow$ & \ Tokens\ $\downarrow$ & Latency\ (s)\ $\downarrow$ & ActRatio\ $\downarrow$ 

\\ \specialrule{0.05em}{0.3em}{0.1em}
\rowcolor[gray]{0.9}
\multicolumn{10}{c}{\textbf{\textit{Qwen2.5-7B Series}}}
\\ \specialrule{0.05em}{0.1em}{0.3em}
\multirow{1}{*}{\makecell[c]{ Original}}
&-       & 92.17$_{\textcolor{Maroon}{\text{0.00}\downarrow}}$               
         & 298.63              
         & 3.83$_{\textcolor{ForestGreen}{\text{1.00}\times}}$                       
         & -               
         & 61.60$_{\textcolor{Maroon}{\text{0.00}\downarrow}}$     
         & 571.64 
         & 6.35$_{\textcolor{ForestGreen}{\text{1.00}\times}}$                  
         & -
\\ \midrule

\multirow{3}{*}{\makecell[c]{Prompt \\ \cite{xia2025tokenskip}}}
&0.9     & 91.47$_{\textcolor{Maroon}{\text{0.70}\downarrow}}$ & 281.65                                  & 3.74$_{\textcolor{ForestGreen}{\text{1.02}\times}}$
         & 0.97    
         & \textcolor{red}{60.40}$_{\textcolor{Maroon}{\text{1.22}\downarrow}}$   
         & 553.27                                     
         & 6.27$_{\textcolor{ForestGreen}{\text{1.01}\times}}$                        
         & 0.97                   \\ 
&0.7     & \textcolor{red}{90.44}$_{\textcolor{Maroon}{\text{1.73}\downarrow}}$                         
         & 287.32                                     
         & 3.77$_{\textcolor{ForestGreen}{\text{1.02}\times}}$
         & 0.99      
         & \textcolor{red}{59.90}$_{\textcolor{Maroon}{\text{1.72}\downarrow}}$
         & 567.49                        
         & 6.34$_{\textcolor{ForestGreen}{\text{1.00}\times}}$                         & 0.99                   \\ 
&0.5     & \textcolor{red}{89.69}$_{\textcolor{Maroon}{\text{2.48}\downarrow}}$          
         & 283.20                                     
         & 3.69$_{\textcolor{ForestGreen}{\text{1.04}\times}}$
         & 0.98           
         & \textcolor{red}{60.30}$_{\textcolor{Maroon}{\text{1.32}\downarrow}}$
         & 577.07                         
         & 6.37$_{\textcolor{ForestGreen}{\text{1.00}\times}}$                              & 1.01                   
\\ \midrule

\multirow{3}{*}{\makecell[c]{Truncation \\ \cite{xia2025tokenskip}}}
&0.9     & 79.68$_{\textcolor{Maroon}{\text{12.49}\downarrow}}$& 272.32                                  & 3.47$_{\textcolor{ForestGreen}{\text{1.10}\times}}$
         & 0.91   
         & 42.70$_{\textcolor{Maroon}{\text{18.92}\downarrow}}$
         & 514.48         
         & 5.72$_{\textcolor{ForestGreen}{\text{1.11}\times}}$                         & 0.90                   \\ 
&0.7     & 52.42$_{\textcolor{Maroon}{\text{39.75}\downarrow}}$& 210.55                                  & 2.68$_{\textcolor{ForestGreen}{\text{1.43}\times}}$ 
         & 0.70           
         & 35.30$_{\textcolor{Maroon}{\text{26.32}\downarrow}}$ 
         & 405.86                         
         & 4.50$_{\textcolor{ForestGreen}{\text{1.41}\times}}$
         & 0.71                   \\ 
&0.5     & 12.30$_{\textcolor{Maroon}{\text{79.87}\downarrow}}$& 154.68                                  & 2.01$_{\textcolor{ForestGreen}{\text{1.91}\times}}$
         & 0.52  
         & 8.50$_{\textcolor{Maroon}{\text{53.12}\downarrow}}$         
         & 285.82                         
         & 3.71$_{\textcolor{ForestGreen}{\text{1.71}\times}}$
         & 0.50                   
\\ \midrule

\multirow{3}{*}{\makecell[c]{CoD \\ \cite{xu2025chain}}}
&0.9     & 90.98$_{\textcolor{Maroon}{\text{1.19}\downarrow}}$ & 257.77                
         & 3.43$_{\textcolor{ForestGreen}{\text{1.12}\times}}$
         & 0.89
         & 59.10$_{\textcolor{Maroon}{\text{2.52}\downarrow}}$                                  & 508.76 
         & 5.66$_{\textcolor{ForestGreen}{\text{1.12}\times}}$
         & 0.89                  \\
&0.7     & 87.54$_{\textcolor{Maroon}{\text{4.63}\downarrow}}$ & 223.02                 
         & 2.97$_{\textcolor{ForestGreen}{\text{1.29}\times}}$
         & 0.77                   
         & 56.60$_{\textcolor{Maroon}{\text{5.02}\downarrow}}$
         & 394.43                 
         & 4.53$_{\textcolor{ForestGreen}{\text{1.40}\times}}$
         & 0.69                  \\
&0.5     & 84.96$_{\textcolor{Maroon}{\text{7.21}\downarrow}}$ & 150.61                 
         & 1.99$_{\textcolor{ForestGreen}{\text{1.92}\times}}$
         & 0.52                   
         & 51.30$_{\textcolor{Maroon}{\text{10.32}\downarrow}}$                                 & 314.40                  
         & 3.61$_{\textcolor{ForestGreen}{\text{1.76}\times}}$
         & 0.55
\\ \midrule

\multirow{3}{*}{\makecell[c]{Tokenskip \\ \cite{xia2025tokenskip}}}
&0.9     & 91.13$_{\textcolor{Maroon}{\text{1.04}\downarrow}}$ & 277.21                                  & 3.62$_{\textcolor{ForestGreen}{\text{1.06}\times}}$
         & 0.93   
         & 58.40$_{\textcolor{Maroon}{\text{3.22}\downarrow}}$ & 525.91         
         & 5.85$_{\textcolor{ForestGreen}{\text{1.09}\times}}$
         & 0.92                   \\ 
&0.7     & 85.69$_{\textcolor{Maroon}{\text{6.48}\downarrow}}$ & 223.83                                  & 2.87$_{\textcolor{ForestGreen}{\text{1.33}\times}}$
         & 0.75           
         & 52.00$_{\textcolor{Maroon}{\text{9.62}\downarrow}}$ & 417.30                         & 4.62$_{\textcolor{ForestGreen}{\text{1.37}\times}}$
         & 0.73                   \\ 
&0.5     & 83.47$_{\textcolor{Maroon}{\text{8.70}\downarrow}}$ & 157.58                                  & 2.07$_{\textcolor{ForestGreen}{\text{1.85}\times}}$
         & 0.53  
         & 47.10$_{\textcolor{Maroon}{\text{14.52}\downarrow}}$
         & 297.25                         
         & 3.30$_{\textcolor{ForestGreen}{\text{1.92}\times}}$
         & 0.52
\\ \midrule

\multirow{3}{*}{\makecell[c]{UCoT}}
&0.9     & \textcolor{red}{91.72}$_{\textcolor{Maroon}{\text{0.45}\downarrow}}$                     
         & \textcolor{red}{244.88}                
         & \textcolor{red}{3.32}$_{\textcolor{ForestGreen}{\text{1.15}\times}}$
         & \textcolor{red}{0.85}  
         & 60.10$_{\textcolor{Maroon}{\text{1.52}\downarrow}}$ & \textcolor{red}{491.61}  
         & \textcolor{red}{5.62}$_{\textcolor{ForestGreen}{\text{1.13}\times}}$
         & \textcolor{red}{0.86}          \\ 
&0.7     & 87.98$_{\textcolor{Maroon}{\text{4.19}\downarrow}}$
         & \textcolor{red}{194.63}                            
         & \textcolor{red}{2.57}$_{\textcolor{ForestGreen}{\text{1.49}\times}}$  
         & \textcolor{red}{0.65}  
         & 58.80$_{\textcolor{Maroon}{\text{2.82}\downarrow}}$ & \textcolor{red}{388.72}               
         & \textcolor{red}{4.37}$_{\textcolor{ForestGreen}{\text{1.45}\times}}$
         & \textcolor{red}{0.68}          \\ 
&0.5     & 86.55$_{\textcolor{Maroon}{\text{5.62}\downarrow}}$    
         & \textcolor{red}{140.36}         
         & \textcolor{red}{1.86}$_{\textcolor{ForestGreen}{\text{2.06}\times}}$ 
         & \textcolor{red}{0.48}  
         & 53.70$_{\textcolor{Maroon}{\text{7.92}\downarrow}}$    
         & \textcolor{red}{280.10}                
         & \textcolor{red}{3.17}$_{\textcolor{ForestGreen}{\text{2.00}\times}}$
         & \textcolor{red}{0.49}
\\ \specialrule{0.05em}{0.3em}{0.1em}

\rowcolor[gray]{0.9}
\multicolumn{10}{c}{\textbf{\textit{Llama3.1-8B Series}}}
\\ \specialrule{0.05em}{0.1em}{0.3em}

\multirow{1}{*}{\makecell[c]{Original}}
&-        & 87.26$_{\textcolor{Maroon}{\text{0.00}\downarrow}}$
          & 212.13
          & 2.44$_{\textcolor{ForestGreen}{\text{1.00}\times}}$
          & -                            
          & 42.70$_{\textcolor{Maroon}{\text{0.00}\downarrow}}$     
          & 574.28
          & 6.88$_{\textcolor{ForestGreen}{\text{1.00}\times}}$
          & -
\\ \midrule

\multirow{3}{*}{\makecell[c]{Prompt \\ \cite{xia2025tokenskip}}}
&0.9     & \textcolor{red}{87.32}$_{\textcolor{Maroon}{\text{0.06}\uparrow}}$
         & 207.89                                     
         & 2.40$_{\textcolor{ForestGreen}{\text{1.02}\times}}$      
         & 0.98  
         & \textcolor{red}{42.60}$_{\textcolor{Maroon}{\text{0.10}\downarrow}}$
         & 528.34         
         & 6.75$_{\textcolor{ForestGreen}{\text{1.02}\times}}$                         & 0.92                   \\ 
&0.7     & \textcolor{red}{86.79}$_{\textcolor{Maroon}{\text{0.47}\downarrow}}$
         & 205.88                                     
         & 2.33$_{\textcolor{ForestGreen}{\text{1.05}\times}}$
         & 0.97           
         & \textcolor{red}{42.30}$_{\textcolor{Maroon}{\text{0.40}\downarrow}}$
         & 534.08                         
         & 6.92$_{\textcolor{ForestGreen}{\text{0.99}\times}}$
         & 0.93                   \\ 
&0.5     & \textcolor{red}{87.16}$_{\textcolor{Maroon}{\text{0.10}\downarrow}}$
         & 189.32                                     
         & 2.15$_{\textcolor{ForestGreen}{\text{1.13}\times}}$
         & 0.89  
         & \textcolor{red}{42.50}$_{\textcolor{Maroon}{\text{0.20}\downarrow}}$             
         & 533.79                        
         & 6.89$_{\textcolor{ForestGreen}{\text{1.00}\times}}$     
         & 0.93                   
\\ \midrule

\multirow{3}{*}{\makecell[c]{Truncation \\ \cite{xia2025tokenskip}}}
&0.9     & 62.55$_{\textcolor{Maroon}{\text{24.71}\downarrow}}$
         & 193.91                                  
         & 2.22$_{\textcolor{ForestGreen}{\text{1.10}\times}}$
         & 0.91   
         & 37.30$_{\textcolor{Maroon}{\text{5.40}\downarrow}}$
         & 511.10         
         & 6.72$_{\textcolor{ForestGreen}{\text{1.02}\times}}$
         & 0.89                   \\ 
&0.7     & 27.48$_{\textcolor{Maroon}{\text{59.78}\downarrow}}$
         & 148.49                                  
         & \textcolor{red}{1.72}$_{\textcolor{ForestGreen}{\text{1.42}\times}}$
         & 0.70           
         & 35.00$_{\textcolor{Maroon}{\text{7.70}\downarrow}}$ 
         & 413.48                         
         & \textcolor{red}{4.52}$_{\textcolor{ForestGreen}{\text{1.52}\times}}$
         & 0.72                   \\ 
&0.5     & 7.32$_{\textcolor{Maroon}{\text{79.94}\downarrow}}$          
         & 108.19                  
         & \textcolor{red}{1.25}$_{\textcolor{ForestGreen}{\text{1.95}\times}}$
         & 0.51  
         & 12.40$_{\textcolor{Maroon}{\text{30.3}\downarrow}}$ & 304.37                        & 3.89$_{\textcolor{ForestGreen}{\text{1.77}\times}}$
         & 0.53                   
\\ \midrule

\multirow{3}{*}{\makecell[c]{CoD \\ \cite{xu2025chain}}}
&0.9     & 85.82$_{\textcolor{Maroon}{\text{1.44}\downarrow}}$                                           & 197.28                
         & 2.28$_{\textcolor{ForestGreen}{\text{1.07}\times}}$
         & 0.93
         & 39.60$_{\textcolor{Maroon}{\text{3.10}\downarrow}}$                                  & 545.57 
         & 6.57$_{\textcolor{ForestGreen}{\text{1.05}\times}}$     
         & 0.95                  \\
&0.7     & 83.90$_{\textcolor{Maroon}{\text{3.36}\downarrow}}$ & 154.85                 
         & 1.84$_{\textcolor{ForestGreen}{\text{1.33}\times}}$
         & 0.73                   
         & 36.70$_{\textcolor{Maroon}{\text{6.00}\downarrow}}$                                  & 413.48                 
         & 4.96$_{\textcolor{ForestGreen}{\text{1.39}\times}}$
         & 0.72                  \\
&0.5     & 73.46$_{\textcolor{Maroon}{\text{13.8}\downarrow}}$ & 110.31                 
         & 1.27$_{\textcolor{ForestGreen}{\text{1.92}\times}}$
         & 0.52                   
         & 29.80$_{\textcolor{Maroon}{\text{12.9}\downarrow}}$                                  & 321.60                  
         & 3.88$_{\textcolor{ForestGreen}{\text{1.77}\times}}$
         & 0.56                  
\\ \midrule

\multirow{3}{*}{\makecell[c]{Tokenskip \\ \cite{xia2025tokenskip}}}
&0.9     & 85.79$_{\textcolor{Maroon}{\text{1.47}\downarrow}}$ & 197.16                                  & 2.37$_{\textcolor{ForestGreen}{\text{1.03}\times}}$ 
         & 0.94   
         & 40.10$_{\textcolor{Maroon}{\text{2.60}\downarrow}}$  & 511.12          
         & 6.15$_{\textcolor{ForestGreen}{\text{1.12}\times}}$
         & 0.89                   \\ 
&0.7     & 82.57$_{\textcolor{Maroon}{\text{4.69}\downarrow}}$ & 161.21                                  & 1.86$_{\textcolor{ForestGreen}{\text{1.31}\times}}$
         & 0.76           
         & 37.90$_{\textcolor{Maroon}{\text{4.80}\downarrow}}$  & 413.48                        & 4.95$_{\textcolor{ForestGreen}{\text{1.39}\times}}$
         & 0.72                   \\ 
&0.5     & 81.80$_{\textcolor{Maroon}{\text{5.46}\downarrow}}$ & 135.76                                  & 1.56$_{\textcolor{ForestGreen}{\text{1.56}\times}}$
         & 0.64  
         & 36.80$_{\textcolor{Maroon}{\text{5.90}\downarrow}}$ & 327.34                        & 4.01$_{\textcolor{ForestGreen}{\text{1.72}\times}}$
         & 0.57
\\ \midrule

\multirow{3}{*}{\makecell[c]{UCoT}}
&0.9     & 86.32$_{\textcolor{Maroon}{\text{0.94}\downarrow}}$ & \textcolor{red}{180.31}                         & \textcolor{red}{2.19}$_{\textcolor{ForestGreen}{\text{1.11}\times}}$
         & \textcolor{red}{0.85}  
         & 40.50$_{\textcolor{Maroon}{\text{2.20}\downarrow}}$  & \textcolor{red}{499.62}
         & \textcolor{red}{6.07}$_{\textcolor{ForestGreen}{\text{1.13}\times}}$
         & \textcolor{red}{0.87}          \\ 
&0.7     & 84.17$_{\textcolor{Maroon}{\text{3.09}\downarrow}}$ & \textcolor{red}{140.01}                         & 1.73$_{\textcolor{ForestGreen}{\text{1.41}\times}}$
         & \textcolor{red}{0.66} 
         & 39.50$_{\textcolor{Maroon}{\text{3.20}\downarrow}}$ & \textcolor{red}{396.25}            
         & 4.80$_{\textcolor{ForestGreen}{\text{1.43}\times}}$              
         & \textcolor{red}{0.69}          \\ 
&0.5     & 83.62$_{\textcolor{Maroon}{\text{3.64}\downarrow}}$  
         & \textcolor{red}{101.82}                
         & 1.26$_{\textcolor{ForestGreen}{\text{1.94}\times}}$
         & \textcolor{red}{0.48} 
         & 38.00$_{\textcolor{Maroon}{\text{4.70}\downarrow}}$ 
         & \textcolor{red}{298.63} 
         & \textcolor{red}{3.79}$_{\textcolor{ForestGreen}{\text{1.82}\times}}$
         & \textcolor{red}{0.52}
\\ \midrule
\bottomrule
\end{tabular}}
\caption{Performance comparison of the CoT compression task under four compression ratio settings. For the accuracy and end-to-end latency, we use color \raisebox{0.5ex}{\colorbox{Maroon}{\rule{-1pt}{0pt}}} and color \raisebox{0.5ex}{\colorbox{ForestGreen}{\rule{-1pt}{0pt}}} respectively to mark the performance changes after compression compared with those of the original model. In addition, we mark the optimal performance indicators under each compression ratio setting with color \raisebox{0.5ex}{\colorbox{Red}{\rule{-1pt}{0pt}}}.}

\label{table:main}
\end{table*}
%%%%%%%%%%% Table %%%%%%%%%%%%%%%%%%%

We compare our method UCoT to four baselines: Prompt \cite{xia2025tokenskip}, Truncation \cite{xia2025tokenskip}, CoD \cite{xu2025chain}, and Tokenskip \cite{xia2025tokenskip} under three different compression ratios (0.9, 0.7, and 0.5) to verify the effectiveness of our method. 
The specific introductions and implementation processes of the comparison methods are in Appendices \ref{A.2} and \ref{A.3}.
At the same time, we conduct a systematic evaluation for the performance of each method in the field of CoT compression from four key metrics: accuracy, the number of compressed CoT tokens, end-to-end latency, and actual compression ratio\footnote{For simplicity, these key metrics are denoted as Acc, Tokens, Latency, and ActRatio in the results tables, respectively.}, which have also been proposed and utilized in prior work \cite{xia2025tokenskip}. 
Among them, accuracy is used to measure the ability of the method to maintain the reasoning performance of reasoning models; end-to-end latency reflects the reasoning efficiency of reasoning frameworks; the number of compressed CoT tokens and the actual compression ratio reflect the execution effect of the method under different compression ratio settings.

%%%%%%%%%%%%% Table-Ablation-Ratio0.7 %%%%%%%%%%%%%
\begin{table*}[h]
\centering
\renewcommand\arraystretch{0.7}
\resizebox{1\textwidth}{!}{
\begin{tabular}{lc|lccc|lccc}
\toprule \midrule
        &  \multicolumn{1}{c}{}  &  \multicolumn{4}{c}{\textbf{GSM8K}}   & \multicolumn{4}{c}{\textbf{MATH-500}} \\ \cmidrule(lr){3-6} \cmidrule(lr){7-10}
\textbf{Methods} & \multicolumn{1}{c}{\textbf{Types}}       & Acc.\ (\%)\ $\uparrow$ & \ Tokens\ $\downarrow$ & \multicolumn{1}{c}{ Latency\ (s)\ $\downarrow$ }& ActRatio\ $\downarrow$ & Acc.\ (\%)\ $\uparrow$ & \ Tokens\ $\downarrow$ & Latency\ (s)\ $\downarrow$ & ActRatio\ $\downarrow$ 

\\ \specialrule{0.05em}{0.3em}{0.1em}
\rowcolor[gray]{0.9}
\multicolumn{10}{c}{\textbf{\textit{Qwen2.5-7B Series}}}
\\ \specialrule{0.05em}{0.1em}{0.3em}
\multirow{1}{*}{\makecell[c]{Original}} 
&-       & 92.17$_{\textcolor{Maroon}{\text{0.00}\downarrow}}$               
         & 298.63              
         & 3.83$_{\textcolor{ForestGreen}{\text{1.00}\times}}$                       
         & -               
         & 61.60$_{\textcolor{Maroon}{\text{0.00}\downarrow}}$     
         & 571.64 
         & 6.35$_{\textcolor{ForestGreen}{\text{1.00}\times}}$                  
         & -
\\ \midrule

\multirow{4}{*}{\makecell[c]{UCoT}}
&with all& \textcolor{red}{87.98$_{\textcolor{Maroon}{\text{4.19}\downarrow}}$}
         & \textcolor{red}{194.63}                            
         & \textcolor{red}{2.57$_{\textcolor{ForestGreen}{\text{1.49}\times}}$}  
         & \textcolor{red}{0.65}  
         & 58.80$_{\textcolor{Maroon}{\text{2.82}\downarrow}}$ 
         & \textcolor{red}{388.72}               
         & \textcolor{red}{4.37$_{\textcolor{ForestGreen}{\text{1.45}\times}}$}
         & \textcolor{red}{0.68}          \\ 
&\textbf{w/o ST}     
         & 73.23$_{\textcolor{Maroon}{\text{18.94}\downarrow}}$
         & 220.99                            
         & 2.97$_{\textcolor{ForestGreen}{\text{1.29}\times}}$  
         & 0.74  
         & 47.50$_{\textcolor{Maroon}{\text{14.12}\downarrow}}$ 
         & 390.84               
         & 4.40$_{\textcolor{ForestGreen}{\text{1.44}\times}}$
         & \textcolor{red}{0.68}          \\ 
&\textbf{w/o $\mathcal{L}_{\text{sem}}$}     
         & 87.32$_{\textcolor{Maroon}{\text{4.85}\downarrow}}$    
         & 274.74         
         & 3.74$_{\textcolor{ForestGreen}{\text{1.02}\times}}$ 
         & 0.92  
         & \textcolor{red}{59.70$_{\textcolor{Maroon}{\text{1.92}\downarrow}}$}    
         & 554.49                
         & 6.32$_{\textcolor{ForestGreen}{\text{1.00}\times}}$
         & 0.97           \\
&\textbf{w/o $\mathcal{R}$}    
         & 71.53$_{\textcolor{Maroon}{\text{15.73}\downarrow}}$    
         & 206.05         
         & 2.86$_{\textcolor{ForestGreen}{\text{1.34}\times}}$ 
         & 0.69  
         & 44.90$_{\textcolor{Maroon}{\text{16.72}\downarrow}}$    
         & 417.30                
         & 4.73$_{\textcolor{ForestGreen}{\text{1.34}\times}}$
         & 0.73
\\ \specialrule{0.05em}{0.3em}{0.1em}

\rowcolor[gray]{0.9}
\multicolumn{10}{c}{\textbf{\textit{Llama3.1-8B Series}}}
\\ \specialrule{0.05em}{0.1em}{0.3em}

\multirow{1}{*}{\makecell[c]{Original}}
&-        & 87.26$_{\textcolor{Maroon}{\text{0.00}\downarrow}}$
          & 212.13
          & 2.44$_{\textcolor{ForestGreen}{\text{1.00}\times}}$
          & -                            
          & 42.70$_{\textcolor{Maroon}{\text{0.00}\downarrow}}$     
          & 574.28
          & 6.88$_{\textcolor{ForestGreen}{\text{1.00}\times}}$
          & -
\\ \midrule

\multirow{4}{*}{\makecell[c]{UCoT}}
&with all          & 84.17$_{\textcolor{Maroon}{\text{3.09}\downarrow}}$ 
         & \textcolor{red}{140.01}                        
         & \textcolor{red}{1.73$_{\textcolor{ForestGreen}{\text{1.42}\times}}$}
         & \textcolor{red}{0.66} 
         & 39.50$_{\textcolor{Maroon}{\text{3.20}\downarrow}}$ 
         & \textcolor{red}{396.25}            
         & \textcolor{red}{4.80$_{\textcolor{ForestGreen}{\text{1.43}\times}}$}      
         & \textcolor{red}{0.69}          \\ 
&\textbf{w/o ST}     & 70.20$_{\textcolor{Maroon}{\text{17.06}\downarrow}}$ 
         & 150.61                         
         & 1.89$_{\textcolor{ForestGreen}{\text{1.29}\times}}$
         & 0.71 
         & 25.70$_{\textcolor{Maroon}{\text{17.00}\downarrow}}$ 
         & 407.74            
         & 4.87$_{\textcolor{ForestGreen}{\text{1.41}\times}}$              
         & 0.71          \\ 
&\textbf{w/o $\mathcal{L}_{\text{sem}}$}     & \textcolor{red}{86.49$_{\textcolor{Maroon}{\text{0.77}\downarrow}}$}  
         & 199.40                
         & 2.37$_{\textcolor{ForestGreen}{\text{1.03}\times}}$
         & 0.94 
         & \textcolor{red}{40.30$_{\textcolor{Maroon}{\text{2.40}\downarrow}}$} 
         & 522.59 
         & 6.42$_{\textcolor{ForestGreen}{\text{1.07}\times}}$
         & 0.91           \\
&\textbf{w/o $\mathcal{R}$}     & 67.58$_{\textcolor{Maroon}{\text{19.68}\downarrow}}$  
         & 144.25                
         & 1.80$_{\textcolor{ForestGreen}{\text{1.41}\times}}$
         & 0.68 
         & 22.80$_{\textcolor{Maroon}{\text{19.90}\downarrow}}$ 
         & 389.49 
         & 4.82$_{\textcolor{ForestGreen}{\text{1.43}\times}}$
         & \textcolor{red}{0.69}
\\ \midrule
\bottomrule
\end{tabular}}
\caption{Performance comparison of ablation study.}

\label{tab:ablation}
\end{table*}
%%%%%%%%%%% Table-Ablation-Ratio0.7 %%%%%%%%%%%%%%%%%%%

%%%%%%%%%%%%% Supplementary Table2 %%%%%%%%%%%%%
\begin{table*}[!h]
\centering
\renewcommand\arraystretch{0.7}
\resizebox{1\textwidth}{!}{
\begin{tabular}{cc|lccc|lccc}
\toprule \midrule
        &  \multicolumn{1}{c}{}  &  \multicolumn{4}{c}{\textbf{GPQA}}   & \multicolumn{4}{c}{\textbf{HumanEval}} \\ \cmidrule(lr){3-6} \cmidrule(lr){7-10}
\textbf{Methods} & \multicolumn{1}{c}{\textbf{Types}}       & Acc.\ (\%)\ $\uparrow$ & \ Tokens\ $\downarrow$ & \multicolumn{1}{c}{ Latency\ (s)\ $\downarrow$ }& ActRatio\ $\downarrow$ & Acc.\ (\%)\ $\uparrow$ & \ Tokens\ $\downarrow$ & Latency\ (s)\ $\downarrow$ & ActRatio\ $\downarrow$ 

\\ \specialrule{0.05em}{0.3em}{0.1em}
\rowcolor[gray]{0.9}
\multicolumn{10}{c}{\textbf{\textit{Qwen3-8B}}}
\\ \specialrule{0.05em}{0.1em}{0.3em}
\multirow{1}{*}{\makecell[c]{ Original}}
&-       & 60.32$_{\textcolor{Maroon}{\text{0.00}\downarrow}}$               
         & 8252.18              
         & 542.70$_{\textcolor{ForestGreen}{\text{1.00}\times}}$                       
         & -               
         & 51.83$_{\textcolor{Maroon}{\text{0.00}\downarrow}}$     
         & 1967.30 
         & 142.67$_{\textcolor{ForestGreen}{\text{1.00}\times}}$                  
         & -
\\ \midrule

\multirow{3}{*}{\makecell[c]{Tokenskip \\ \cite{xia2025tokenskip}}}
&0.9     & 58.72$_{\textcolor{Maroon}{\text{1.60}\downarrow}}$ & 7683.74                        & 504.61$_{\textcolor{ForestGreen}{\text{1.08}\times}}$
         & 0.93   
         & \textcolor{red}{48.75}$_{\textcolor{Maroon}{\text{3.08}\downarrow}}$
         & 1845.66         
         & 144.83$_{\textcolor{ForestGreen}{\text{0.99}\times}}$
         & 0.93                   \\ 
&0.7     & 57.79$_{\textcolor{Maroon}{\text{2.53}\downarrow}}$ & 6213.35                        & 408.25$_{\textcolor{ForestGreen}{\text{1.33}\times}}$
         & 0.75           
         & 46.13$_{\textcolor{Maroon}{\text{5.70}\downarrow}}$ & \textcolor{red}{1462.13}
         & \textcolor{red}{108.91}$_{\textcolor{ForestGreen}{\text{1.31}\times}}$
         & \textcolor{red}{0.73}                   \\ 
&0.5     & 54.23$_{\textcolor{Maroon}{\text{6.09}\downarrow}}$ & 4388.54                        & 294.63$_{\textcolor{ForestGreen}{\text{1.84}\times}}$
         & 0.53  
         & 42.52$_{\textcolor{Maroon}{\text{9.31}\downarrow}}$
         & 1025.00                         
         & \textcolor{red}{78.18}$_{\textcolor{ForestGreen}{\text{1.82}\times}}$
         & \textcolor{red}{0.52}
\\ \midrule

\multirow{3}{*}{\makecell[c]{UCoT}}
&0.9     & \textcolor{red}{59.30}$_{\textcolor{Maroon}{\text{1.02}\downarrow}}$        
         & \textcolor{red}{7097.48}                
         & \textcolor{red}{469.22}$_{\textcolor{ForestGreen}{\text{1.16}\times}}$
         & \textcolor{red}{0.86}  
         & 48.33$_{\textcolor{Maroon}{\text{3.50}\downarrow}}$ 
         & \textcolor{red}{1761.89}  
         & \textcolor{red}{126.63}$_{\textcolor{ForestGreen}{\text{1.13}\times}}$
         & \textcolor{red}{0.89}          \\ 
&0.7     & \textcolor{red}{58.27}$_{\textcolor{Maroon}{\text{2.05}\downarrow}}$
         & \textcolor{red}{5647.24}                            
         & \textcolor{red}{374.03}$_{\textcolor{ForestGreen}{\text{1.45}\times}}$  
         & \textcolor{red}{0.68}  
         & \textcolor{red}{48.12}$_{\textcolor{Maroon}{\text{3.71}\downarrow}}$ 
         & 1523.82               
         & 112.86$_{\textcolor{ForestGreen}{\text{1.26}\times}}$
         & 0.77          \\ 
&0.5     & \textcolor{red}{56.86}$_{\textcolor{Maroon}{\text{3.46}\downarrow}}$    
         & \textcolor{red}{4065.58}         
         & \textcolor{red}{277.23}$_{\textcolor{ForestGreen}{\text{1.96}\times}}$ 
         & \textcolor{red}{0.49}  
         & \textcolor{red}{46.55}$_{\textcolor{Maroon}{\text{5.28}\downarrow}}$    
         & \textcolor{red}{1021.93}                
         & 78.84$_{\textcolor{ForestGreen}{\text{1.81}\times}}$
         & \textcolor{red}{0.52}
\\ \specialrule{0.05em}{0.3em}{0.1em}

\rowcolor[gray]{0.9}
\multicolumn{10}{c}{\textbf{\textit{Deepseek-R1-Distill-Qwen-7B}}}
\\ \specialrule{0.05em}{0.1em}{0.3em}

\multirow{1}{*}{\makecell[c]{Original}}
&-        & 55.53$_{\textcolor{Maroon}{\text{0.00}\downarrow}}$
          & 9076.19
          & 323.63$_{\textcolor{ForestGreen}{\text{1.00}\times}}$
          & -                            
          & 49.16$_{\textcolor{Maroon}{\text{0.00}\downarrow}}$     
          & 1765.04
          & 46.83$_{\textcolor{ForestGreen}{\text{1.00}\times}}$
          & -
\\ \midrule

\multirow{3}{*}{\makecell[c]{Tokenskip \\ \cite{xia2025tokenskip}}}
&0.9     & 53.79$_{\textcolor{Maroon}{\text{1.74}\downarrow}}$ & 8347.48                        & 300.96$_{\textcolor{ForestGreen}{\text{1.08}\times}}$ 
         & 0.92   
         & 47.32$_{\textcolor{Maroon}{\text{1.84}\downarrow}}$  & 1589.36          
         & 45.92$_{\textcolor{ForestGreen}{\text{1.02}\times}}$
         & 0.90                   \\ 
&0.7     & 52.88$_{\textcolor{Maroon}{\text{2.65}\downarrow}}$ & 6626.87                        & 242.99$_{\textcolor{ForestGreen}{\text{1.33}\times}}$
         & 0.73           
         & 44.03$_{\textcolor{Maroon}{\text{5.13}\downarrow}}$  & 1275.88              & 39.36$_{\textcolor{ForestGreen}{\text{1.19}\times}}$
         & 0.72                   \\ 
&0.5     & 51.50$_{\textcolor{Maroon}{\text{4.03}\downarrow}}$ & 5247.19                        & 190.54$_{\textcolor{ForestGreen}{\text{1.70}\times}}$
         & 0.58  
         & 43.79$_{\textcolor{Maroon}{\text{5.37}\downarrow}}$ & 900.040               & 27.88$_{\textcolor{ForestGreen}{\text{1.68}\times}}$
         & 0.51
\\ \midrule

\multirow{3}{*}{\makecell[c]{UCoT}}
&0.9     & \textcolor{red}{54.62}$_{\textcolor{Maroon}{\text{0.91}\downarrow}}$ 
         & \textcolor{red}{7624.00}
         & \textcolor{red}{275.92}$_{\textcolor{ForestGreen}{\text{1.17}\times}}$
         & \textcolor{red}{0.84}  
         & \textcolor{red}{48.94}$_{\textcolor{Maroon}{\text{0.22}\downarrow}}$  & \textcolor{red}{1542.48}
         & \textcolor{red}{43.56}$_{\textcolor{ForestGreen}{\text{1.08}\times}}$
         & \textcolor{red}{0.87}          \\ 
&0.7     & \textcolor{red}{52.97}$_{\textcolor{Maroon}{\text{2.56}\downarrow}}$ 
         & \textcolor{red}{6361.32}
         & \textcolor{red}{232.54}$_{\textcolor{ForestGreen}{\text{1.39}\times}}$
         & \textcolor{red}{0.70} 
         & \textcolor{red}{45.76}$_{\textcolor{Maroon}{\text{3.40}\downarrow}}$ 
         & \textcolor{red}{1227.76}            
         & \textcolor{red}{37.72}$_{\textcolor{ForestGreen}{\text{1.24}\times}}$     
         & \textcolor{red}{0.69}          \\ 
&0.5     & \textcolor{red}{52.16}$_{\textcolor{Maroon}{\text{3.37}\downarrow}}$  
         & \textcolor{red}{4635.69}                
         & \textcolor{red}{172.13}$_{\textcolor{ForestGreen}{\text{1.88}\times}}$
         & \textcolor{red}{0.51} 
         & \textcolor{red}{43.96}$_{\textcolor{Maroon}{\text{5.20}\downarrow}}$ 
         & \textcolor{red}{870.68} 
         & \textcolor{red}{25.12}$_{\textcolor{ForestGreen}{\text{1.86}\times}}$
         & \textcolor{red}{0.49}
\\ \midrule
\bottomrule
\end{tabular}}
\caption{Performance comparison of extra long CoT. }

\label{table:supplementary_main_2}
\end{table*}
%%%%%%%%%%% Supplementary Table2 %%%%%%%%%%%%%%%%%%%
\subsection{Performance Comparison} \label{main}
We report the experiment results in \tabref{table:main}.
We find that although current methods struggle to maintain performance with shorter reasoning paths, UCoT finds a balance between effectiveness and efficiency.
% As shown in \tabref{table:main}, in this study, UCoT is systematically compared with four mainstream CoT compression methods under three compression ratio settings of 0.9, 0.7, and 0.5. Through a detailed analysis of four core indicators, it is found that UCoT can effectively achieve CoT compression while maintaining the reasoning performance and efficiency of the original executor relatively well.
Specifically, the Prompt method fails to faithfully compress the CoT to the specified compression ratio, leading to good performance on accuracy using a similar length with the original CoT.
% although the prompt method can preserve the reasoning performance of the executor, it has significant deficiencies in compression effect. When using Qwen2.5-7B as the executor and testing the Math-500 dataset with a compression ratio of 0.5, the actual compression ratio of this method even exceeds 1, indicating that it is difficult to achieve effective compression. 
The Truncation method exhibits the opposite characteristics. 
Although it achieves the given compression rate, the substantial information loss during direct truncation causes a significant drop in model performance.
CoD and Tokenskip are stronger competitors, however, they are still inferior in terms of performance and efficiency to our method.
For example, when the compression ratio is set to 0.5, UCoT reduces the average CoT length of the model by 50.75\%, while the average model accuracy is improved by 5.63\% and 3.23\% compared with CoD and Tokenskip, respectively.
%For example, when the compression ratio is set to 0.5, UCoT only causes the average acc of the model to drop by 5.43\%, which is better than CoD's 11.06\% and Tokenskip's 8.66\%.
%For example, when the compression ratio is set to 0.5, the average model acc of UCoT is improved by 5.63\% and 3.23\% compared with CoD and Tokenskip respectively.
This is because CoD and Tokenskip impair the model's reasoning by imposing rigid output constraints or fine-tuning on external short CoT datasets without information enrichment. 
In contrast, UCoT provides high-quality contextual soft tokens at the input stage, naturally guiding executor to reduce generation by leveraging the reasoning priors. 
This demonstrates the superiority of our compressor-executor framework: the system achieves optimal efficiency not by limiting the model, but by post-reasoning paradigm where the compressor provides the contextual CoT and the executor focuses on the result.
In Appendices \ref{C.1}--\ref{C.3} and \ref{C.7}--\ref{C.8}, we further analyze the impacts of compressor type, executor scale and prompt robustness, alongside investigations into UCoT case studies.

\subsection{Ablation Study}\label{4.2}

To comprehensively investigate the influence of individual components within UCoT on its final performance, we conduct ablation experiments for soft tokens, semantic loss $\mathcal{L}_{\text{sem}}$, and reward factor $\mathcal{R}$ under the compression ratio of 0.7.
We remove each part separately during the executor training, specifically: (1) \textbf{w/o. ST} removes soft tokens from the input to the executor. (2) \textbf{w/o. $\mathcal{L}_{\text{sem}}$} trains the executor without semantic loss. (3) \textbf{w/o. $\mathcal{R}$} removes the reward factor $\mathcal{R}$ from $\mathcal{L}_e$.

As shown in \tabref{tab:ablation}, incorporating soft tokens as part of the input significantly improved the reasoning performance of executor during training. 
This indicates that the contextual CoT output by the compressor can provide effective reasoning guidance for executor.
Moreover, compared to the complete version, \textbf{w/o. $\mathcal{L}_{\text{sem}}$} achieve higher reasoning performance for executor in some settings, but it cannot effectively complete the compression task for CoT, which is similar to the research results of \cite{xu2025softcot}.
This reveals our speculation in Section~\ref{section:3.2} about the importance of guiding large language models to fully utilize soft tokens.
Furthermore, the reward factor $\mathcal{R}$ further enhances the accuracy of the executor, which indicates that it effectively promotes the retention of the original executor's reasoning ability.

\subsection{Applicability to Extra Long CoT}\label{4.3}
Current large reasoning models are capable of generating reasoning paths spanning thousands of tokens to tackle complex tasks, which poses challenges to the effectiveness of CoT compression methods.
To evaluate whether UCoT still performs well in compressing extra-long CoT, we test it on Qwen3-8B \cite{yang2025qwen3} and Deepseek-R1-Distill-Qwen-7B \cite{guo2025deepseek} on two representative datasets for cover complex commonsense reasoning and code generation: GPQA \cite{rein2024gpqa} and HumanEval \cite{chen2021codex}.

Experimental results in \tabref{table:supplementary_main_2}  show that even when dealing with complex data and in cases where R1 models adopt ultra-long CoTs, UCoT can still significantly improve their inference efficiency. For instance, when UCoT is applied to the Deepseek-R1-Distill-Qwen-7B model, the usage of inference tokens on the HumanEval dataset is reduced by 50.67\%, while the performance is improved by 0.17\% compared with Tokenskip, the current SOTA method. Similar improvements are observed in the Qwen3-8B model: UCoT reduces the model's inference token usage on the dataset by 48.05\% and improves its performance by 4.03\% compared with SOTA method Tokenskip.
This highlights UCoT’s ability to balance efficiency and performance.
Additional evaluations on AIME2024 \cite{finkelstein2024artificial} and ASDiv \cite{miao-etal-2020-diverse}, covering mathematical reasoning tasks, are presented in Appendix \ref{C.9}.

\subsection{Effectiveness of Contextual CoT}\label{4.4}
We look into the reasons for the effectiveness of soft tokens from the perspective of information volume with two metrics proposed in \citet{kuratov2025cramming}, namely \textbf{Token Gain} and \textbf{Information Gain}.
% explore the association between the information volume and the model performance in this section to provide a more comprehensive view of the function of UT.
% In this work, we train a small model as a compressor to compress long CoT into continuous short embeddings UT.
% We posit that the UT containing high-quality reasoning information can assist the subsequent reasoning of the executor and reduce the need for the executor to output a long CoT by itself.
% To verify this conjecture, in addition to the main experiment in \tabref{table:main}, we further conducted an exploratory study.
% We quantify the information contained in UT with two metrics proposed in \citet{2025arXiv250213063K}, namely \textbf{Token Gain} and \textbf{Information Gain}.
% These two metrics compute the compression of text information volume in continuous embeddings. 
% Previous work \cite{2025arXiv250213063K} propose two metrics, namely \textbf{Token Gain} and \textbf{Information Gain}, for computing the compression of text information volume in continuous embeddings. 
Specifically, Token Gain describes the difference in th number of tokens that can be correctly decoded with and without the presence of continuous embeddings.
% in the presence of continuous embeddings, compared to the baseline performance of a language model without them. 
In UCoT, this can be described as:
\begin{align}\label{15} 
\mathcal{C}_{tokens} =\mathcal{C}_{tokens}^{\mathcal{M}_{c}+H_{n}}-\mathcal{C}_{tokens}^{\mathcal{M}_{c}},
\end{align}
where $\mathcal{C}_{tokens}^{\mathcal{M}_{c}}$ and $\mathcal{C}_{tokens}^{\mathcal{M}_{c}+H_{n}}$ represent the number of tokens that can be correctly decoded by $\mathcal{M}_{c}$ before and after adding soft tokens $H_{n}$. 
% From the perspective of information theory, this metric quantifies the number of discre tetoken information that the UT adds to the decoding process. 
Information Gain measures the change in Cross-Entropy of the target text by calculating the output uncertainty eliminated by the language model before and after the addition of soft tokens $H_{n}$. 
In our work, Information Gain can be defined as:
\begin{align}\label{16} 
\mathcal{H}_{ce} =\mathcal{H}_{ce}^{\mathcal{M}_{c}}-\mathcal{H}_{ce}^{\mathcal{M}_{c}+H_{n}},
\end{align}
where $\mathcal{H}_{ce}^{\mathcal{M}_{c}+H_{n}}$ and $\mathcal{H}_{ce}^{\mathcal{M}_{c}}$ represent the Cross-Entropy value before and after adding $H_{n}$.

%%%%%%%%%%%%% Table-Interpretability-of-Information %%%%%%%%%%%%%
\begin{table}[h]
\centering
\resizebox{\linewidth}{!}{
\begin{tabular}{cccccc}
\toprule \midrule
 \textbf{Metrics} & \textbf{Values} & \textbf{Acc} & \textbf{Tokens} & \textbf{Latency} & \textbf{Latency} \\
\specialrule{0.05em}{0.3em}{0.1em}
 Original  & -
         & 92.17$_{\textcolor{Maroon}{\text{0.00}\downarrow}}$               
         & 298.63              
         & 3.83$_{\textcolor{ForestGreen}{\text{1.00}\times}}$                       
         & - \\
\specialrule{0.05em}{0.3em}{0.1em}
\multirow{3}{*}{\makecell[c]{Token\\ Gain$\uparrow$}}  &72.35
         & 70.68$_{\textcolor{Maroon}{\text{21.49}\downarrow}}$
         & 218.00                            
         & 2.83$_{\textcolor{ForestGreen}{\text{1.35}\times}}$  
         & 0.73\\ 
& 123.62
         & 81.45$_{\textcolor{Maroon}{\text{10.72}\downarrow}}$
         & 200.08                            
         & 2.62$_{\textcolor{ForestGreen}{\text{1.46}\times}}$  
         & 0.67 \\
& 182.45
         & 86.55$_{\textcolor{Maroon}{\text{5.62}\downarrow}}$    
         & 200.39         
         & 2.61$_{\textcolor{ForestGreen}{\text{1.47}\times}}$ 
         & 0.67 \\
\specialrule{0.05em}{0.3em}{0.1em}
 \multirow{3}{*}{\makecell[c]{Information \\Gain$\uparrow$}}  & 841.07
         & 67.38$_{\textcolor{Maroon}{\text{24.79}\downarrow}}$
         & 212.13                            
         & 2.83$_{\textcolor{ForestGreen}{\text{1.35}\times}}$  
         & 0.71 \\ 
& 1077.82
         & 75.95$_{\textcolor{Maroon}{\text{16.22}\downarrow}}$
         & 206.05                            
         & 2.72$_{\textcolor{ForestGreen}{\text{1.41}\times}}$  
         & 0.69 \\
& 1863.45
         & 85.55$_{\textcolor{Maroon}{\text{6.62}\downarrow}}$    
         & 197.10         
         & 2.65$_{\textcolor{ForestGreen}{\text{1.45}\times}}$ 
         & 0.66 \\
\bottomrule \midrule
\end{tabular}
}
\caption{Token gain and information gain of UCoT.} 
\label{tab:TG_IG}
\vspace{-0.6em}
\end{table}
%%%%%%%%%%% Table-Interpretability-of-Information %%%%%%%%%%%%%%%%%%%

%%%%%%%%%%%%%%%%% Figure-SoftCoT-length %%%%%%%%%%
\begin{figure}[!t]
    \centering
    \includegraphics[width=0.46\textwidth]{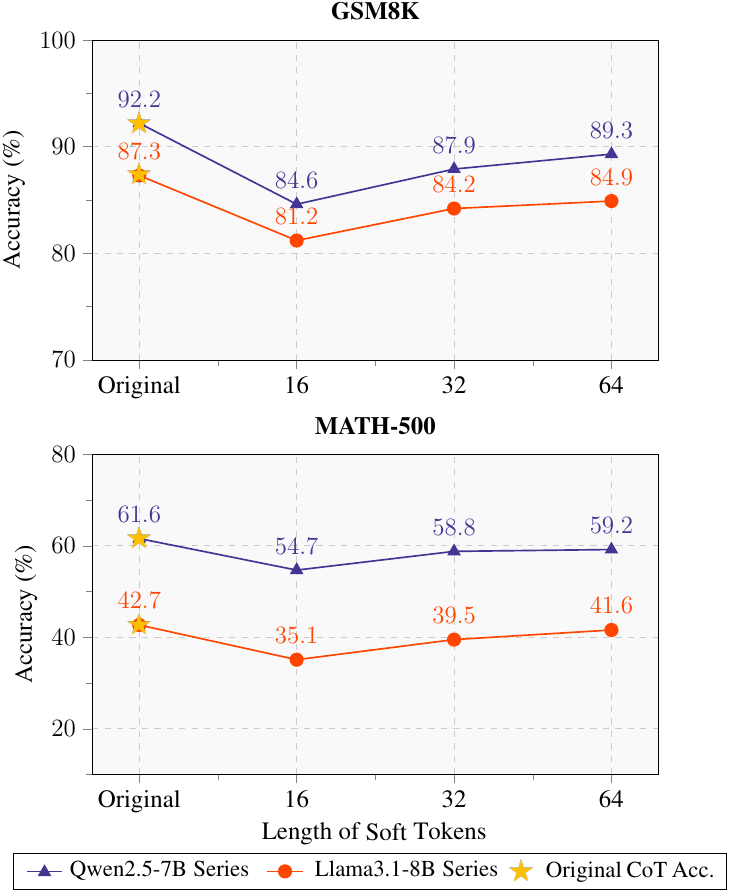}
    \caption{Performance comparison between original CoT and soft tokens of different lengths.}
    \label{fig:length}
\end{figure}
%%%%%%%%%%%%%%%%% Figure-SoftCoT-length %%%%%%%%%%

We control the information in soft tokens by early stopping compressor training at different step, then measure both the information volume of contextual CoT and UCoT performance.
The results with the Qwen2.5-7b-Instrust model on the GSM8K dataset are reported in \tabref{tab:TG_IG}.
% We demonstrate the experimental results of the Qwen2.5-7b-Instrust model on the GSM8K dataset with a compression ratio setting of 0.7, as shown in \tabref{tab:TG_IG}.
The positive correlation between the information volume in soft tokens and the performance of UCoT suggests that our method compresses rich reasoning information into the embedding space of LLMs, allowing the model to maintain performance with a shorter reasoning path.
% the information contains in UT and the performance of UCoT are positively correlated.
% Sufficient information in the model input helps improve the response quality.
% Our method compresses rich reasoning information within the embedding space of LLMs, leveraging the expressiveness of high-dimensional representations. This enables efficient transfer of reasoning signals from the compressor to the executor with minimal information loss.
% In addition, we conduct a case study on the soft tokens decoded text in Appendix \ref{C.7}.

\subsection{Analysis of Length of  Contextual CoT}
Recent literature on prompt compression suggests that the information density of compressed soft tokens is highly sensitive to their allocated length \cite{mu2023learning}.
We explore the association between the length of soft tokens and the model performance in this section.
As shown in \figref{fig:length} the results indicate that as the length of the soft tokens gradually increases, the accuracy of the executor on various test datasets presents a significant upward trend.
Using Qwen2.5-7B-Instruct as the executor and inferencing the GSM8K dataset, the model's accuracy improved by 4.75\% when the soft tokens length increased from 16 to 64, gradually approaching that of the original CoT reasoning.
This result fully demonstrates that as the length of the soft tokens increases, the compressor can provide more complete and accurate reasoning information for the executor.

\subsection{Comparison with Other Latent Reasoning Methods}
In this section, we present a comparative analysis of UCoT and existing latent reasoning methods. Specifically, latent reasoning methods for large models can be categorized into two types: one is the continuous token method, which converts CoT into continuous embedding vectors \cite{hao2024training}; the other is the loop layer method, which enables inputs to flow repeatedly through the same layer within the model 
\cite{geiping2025scaling}.
%%%%%%%%%%%%% Supplementary Table3 %%%%%%%%%%%%%
\begin{table}[h]
\centering
\resizebox{1\linewidth}{!}{
\begin{tabular}{ccccc}
\toprule \midrule
 \textbf{Methods} & \textbf{AIME 2024} & \textbf{GPQA} & \textbf{HumanEval} & \textbf{ASDiv}\\
\specialrule{0.05em}{0.3em}{0.1em}
 Original  & 66.67$_{\textcolor{Maroon}{\text{0.00}\downarrow}}$
        & 60.32$_{\textcolor{Maroon}{\text{0.00}\downarrow}}$ 
        & 51.83$_{\textcolor{Maroon}{\text{0.00}\downarrow}}$ 
        & 93.74$_{\textcolor{Maroon}{\text{0.00}\downarrow}}$ 
          \\
\specialrule{0.05em}{0.3em}{0.1em}
\multirow{1}{*}{\makecell[c]{Depth-Recurrent}}  & 50.00$_{\textcolor{Maroon}{\text{16.7}\downarrow}}$
        & 51.32$_{\textcolor{Maroon}{\text{9.00}\downarrow}}$ 
        & 34.36$_{\textcolor{Maroon}{\text{17.5}\downarrow}}$ 
        & 87.31$_{\textcolor{Maroon}{\text{6.43}\downarrow}}$  
         \\ 
\multirow{1}{*}{\makecell[c]{Coconut}}  & 56.67$_{\textcolor{Maroon}{\text{10.0}\downarrow}}$
        & 55.42$_{\textcolor{Maroon}{\text{4.90}\downarrow}}$ 
        & 43.77$_{\textcolor{Maroon}{\text{8.06}\downarrow}}$ 
        & 88.26$_{\textcolor{Maroon}{\text{5.48}\downarrow}}$   
         \\ 
\multirow{1}{*}{\makecell[c]{LightThinker}}  & 63.33$_{\textcolor{Maroon}{\text{3.34}\downarrow}}$
        & 57.92$_{\textcolor{Maroon}{\text{2.40}\downarrow}}$ 
        & 46.93$_{\textcolor{Maroon}{\text{4.91}\downarrow}}$ 
        & 93.42$_{\textcolor{Maroon}{\text{0.32}\downarrow}}$   
         \\ 
 UCoT-0.9& \textcolor{red}{66.67}$_{\textcolor{Maroon}{\text{0.00}\downarrow}}$        
        & \textcolor{red}{59.30}$_{\textcolor{Maroon}{\text{1.02}\downarrow}}$
        & \textcolor{red}{48.33}$_{\textcolor{Maroon}{\text{3.50}\downarrow}}$
        & \textcolor{red}{93.53}$_{\textcolor{Maroon}{\text{0.21}\downarrow}}$  
         \\ 
UCoT-0.7& 63.33$_{\textcolor{Maroon}{\text{3.34}\downarrow}}$        
        & 58.27$_{\textcolor{Maroon}{\text{2.05}\downarrow}}$
        & 48.12$_{\textcolor{Maroon}{\text{3.71}\downarrow}}$
        & 93.32$_{\textcolor{Maroon}{\text{0.42}\downarrow}}$
         \\
UCoT-0.5& 56.67$_{\textcolor{Maroon}{\text{10.0}\downarrow}}$        
        & 56.86$_{\textcolor{Maroon}{\text{3.46}\downarrow}}$
        & 46.55$_{\textcolor{Maroon}{\text{5.28}\downarrow}}$
        & 92.77$_{\textcolor{Maroon}{\text{0.97}\downarrow}}$   
         \\
\bottomrule\midrule
\end{tabular}
}
\caption{Comparison with other potential reasoning methods with Qwen3-8b backbone.} 
\label{tab:supplementary_table3}
\end{table}
%%%%%%%%%%% Supplementary Table3 %%%%%%%%%%%%%%%%%%%
%To comprehensively evaluate the performance of UCoT, we compared it with current latent reasoning methods. It is important to highlight that although UCoT shares a similar core idea with these methods—embedding reasoning information into the latent space—it differs significantly in both mechanism and objective: UCoT compresses reasoning steps into continuous representations (without recurrent latent reasoning) and shortens the CoT length using pre-generated continuous thoughts (rather than fully replacing discrete CoT).
We set three compression ratios (0.9, 0.7, and 0.5) for UCoT, denoted as UCoT-0.9, UCoT-0.7, and UCoT-0.5 for brevity. 
We then compare UCoT with representative
works of the three types of latent reasoning methods, namely Depth-Recurrent \cite{geiping2025scaling}, Coconut \cite{hao2024training} and LightThinker \cite{zhang-etal-2025-lightthinker}.

The results are presented in \tabref{tab:supplementary_table3}. Experimental findings indicate that under most settings, the performance of Depth-Recurrent and Coconut is at most comparable to that of UCoT-0.5. For instance, on the GPQA dataset, the accuracy rates of UCoT-0.5, Depth-Recurrent, and Coconut are 56.86\%, 51.32\%, and 55.42\%, respectively.
While LightThinker represents a formidable competitor, it nonetheless lags behind our method in maintaining performance stability.
UCoT-0.9 achieves an average improvement of 1.56\% over LightThinker across the four datasets.

\subsection{Case study for Contextual CoT}
In this section, we explore what specific information the soft token contains from original CoT.
We decode the contextual CoT in UCoT into text sequences with the compressor established in Section \ref{section:3.1} to inspect the exact information it provides, and present examples in Figure \ref{fig:case_ut_1} and Figure \ref{fig:case_ut_2} in  Appendix \ref{C.7}.
Such an analysis allows us to verify whether the compressor successfully distills essential reasoning logic into the compressed representations.
We find the text decoded from the soft tokens exhibits reasonable logical structure that closely mirrors the original CoT.

For instance, in the case of Math-500, they both follow a similar reasoning pattern: \textit{fractional quantity derivation $\rightarrow$ total amount aggregation $\rightarrow$ remainder calculation}, with only specific numbers and objects varied.
These structural cues within the soft tokens act as a cognitive template, guiding the executor to maintain logical consistency.
This observation suggests that compressed contextual CoT in UCoT provides demonstration for how to organize reasoning, echoing the principles of in-context learning and recent findings on the importance of reasoning structure \cite{2025arXiv250207374L}.

\section{Conclusion}
In this work, we propose post-reasoning, a paradigm which enables shorter LLM outputs by providing CoT as context rather than generation.
We investigate the factors that affect the model behavior in post-reasoning, and find the latency in contextual CoT generation and the quality of contextual CoT play critical roles in the efficiency and performance of post-reasoning. 
To achieve optimal efficiency-performance trade-off, we further introduce UCoT inspired by the properties of post-reasoning to enable efficient and informative contextual CoT generation.
This method trains a lightweight language model as the compressor to produce compact contextual CoT in the form of soft tokens, preserving essential reasoning structure within a limited length.
The resulting contextual CoT is then utilized by a large executor model to generate accurate response with reduced output length. 
Experiments demonstrate that our method outperforms existing comparative CoT compression methods across a range of compression ratios, and further analysis of the contextual CoT sheds light on the underlying mechanism by which UCoT preserves reasoning capability.

% explore the possibility of using small models to provide CoT for large models. We find that employing CoT outputs from small models effectively reduces the output length of LLMs while maintaining certain performance levels. 
% Inspired by this novel discovery, we propose a cooperative framework for CoT compression, \ie UCoT. 
% This method leverages smaller language models as the compressor to compress the preliminary reasoning step into continuous embeddings (Upfront Thought, UT).
% The embeddings are utilized by a large model to generate accurate response with shorter reasoning paths. 
% To achieve this, we design a two-stage training process. 
% The compressor is trained to generate UT containing similar information to the reasoning path of the original model.
% The executor is rewarded for giving the correct answers based on the UT. 
% Extensive experiments demonstrate that our method outperforms existing comparative CoT compression methods under various compression ratios.
\section*{Limitations}\label{Limitations}
% We primarily focus on the performance in the field of CoT compression in this work. 
% Nevertheless, our research has some limitations in practical application scenarios. 
% First, our approach currently covers only a limited variety of open-source inference models and cannot be directly applied to closed-source models. 
% Our future research plans to expand the experimental scope by incorporating more diverse types of open-source inference models and simultaneously delve into the application possibilities for closed-source models.  
% Second, the current method cannot adaptively determine whether to perform compression or adjust the compression ratio for individual problems. In the future, we will conduct in-depth research on adaptive CoT compression methods.
% Finally, due to computational constraints, we have not conducted experiments on larger LLMs (\eg Qwen2.5-32b Series and Qwen2.5-72b Series). 
% In the future, we will explore the application of UCoT in larger models if possible.

Despite the promising results achieved by UCoT in the context of CoT compression, there are three primary limitations to this study. \textbf{First}, our method has only been evaluated on a limited set of open-source inference models, and its applicability to closed-source models remains unexplored. \textbf{Second}, the method applies a uniform compression ratio across all inputs and lacks the capability to adaptively adjust compression strategies based on the complexity or difficulty of individual reasoning problems. \textbf{Third}, due to computational resource constraints, we have not evaluated UCoT on larger-scale LLMs (\eg Qwen2.5-32B series and Qwen2.5-72B series). While our results are promising, evaluating UCoT on larger-scale models remains an important next step. This would offer a more comprehensive validation of our method's scalabiliy in real-world scenarios.
\section*{Ethics Statement}
UCoT aims to help improve the reasoning efficiency of existing LLMs. 
Our goal is to develop a general reasoning method that can be applied to current mainstream LLMs and to inspire progress towards more efficient LLM reasoning frameworks. 
We explicitly do not encourage any malicious use of our work, especially attempts to circumvent or compromise LLM systems. 
The artifacts and datasets in our work are all under the restriction of the license and follow the intended use.
\section*{Acknowledgment}
We thank all the reviewers and the area chair for their helpful feedback, which aided us in greatly improving the paper.
This work is supported by National Key R\&D Program (2023YFB3107400), National Natural Science Foundation of China (62272371, 62103323, (62521002, U24B20185, U2441240, 62132011), Initiative Postdocs Supporting Program (BX20190275, BX20200270), China Postdoctoral Science Foundation (2019M663723, 2021M692565), and Fundamental Research Funds for the Central Universities under grant (xzy012024144, xzy012025043). Cong Wang was supported in part by Hong Kong Research Grants Council (RGC) under Grants C6015-23G and CRS\_HKUST601/24. Thanks to the New Cornerstone Science Foundation, the Xplorer Prize and K. C. Wong Education Foundation.

\bibliography{custom}
\clearpage
\newpage
\appendix
\section*{Appendix}
%The appendices of this paper are composed of two parts: Appendix \ref{A} focuses on presenting the experimental details, Appendix \ref{B} introduces definitions of Chain-of-Thought Reasoning and Chain-of-Thought Compression, and Appendix \ref{C} emphasizes the analysis of performance expansion.
This paper’s appendices have three parts: Appendix \ref{A} (experimental details), Appendix \ref{B} (CoT compression definitions), and Appendix \ref{C} (performance expansion analysis).

\section{Experiment Setting Details}\label{A}
\subsection{Datasets}\label{A.1}
Table~\ref{tab:datasets} provides an overview of the datasets used in this work, including their domain categories and the sizes of their training and test splits. These datasets span three major task types: mathematical reasoning (GSM8K, MATH, AIME, ASDiv), scientific question answering (GPQA), and code generation (HumanEval). Following standard evaluation protocols from prior work \cite{xia2025tokenskip}, we adopt task-specific settings for each dataset. For the MATH dataset in particular, we report results on the MATH-500 subset to reduce computational overhead, as it has been shown to serve as a reliable proxy for the full test set \cite{xia2025tokenskip}.

\begin{table}[!h]
\centering
\renewcommand\arraystretch{1.2} % 增加行高，改善垂直间距
\setlength{\tabcolsep}{5pt}    % 加宽列间距，改善水平空间
\begin{tabular}{c c c c} 
\toprule
\textbf{Datasets} & \textbf{Train} & \textbf{Test} & \textbf{Domain} \\
\midrule
GSM8K     & 7473  & 1319 & Math Reasoning \\
Math      & 7500  &   500 & Math Reasoning \\
AIME 2024      & -  &   30 & Math Reasoning \\
ASDiv     & -  &   2305 & Math Reasoning \\
GPQA      &   -    &   448 & Scientific QA \\
HumanEval &   -    &   164 & Code Generation \\
\bottomrule
\end{tabular}
\caption{Datasets in the main experiments.}
\label{tab:datasets}
\end{table}

\vspace{-0.3cm}
\subsection{Baselines}\label{A.2}
We conduct extensive experiments comparing four major competitors, including representative CoT compression methods based on discrete prompts and compressed datasets.

\noindent
\textbf{Prompt \cite{xia2025tokenskip}:} 
In this method, we use prompts to guide the LLM to reduce a fixed percentage of the output token in the CoT process. For example, append a piece of text to the sample, "Please reduce your thought process by 50 percent." as input to the sample. 
Compared with CoT compression methods based on discrete prompts that require a large number of discrete prompts to be designed or use small-shot learning, this method only needs a single text instruction and does not need to rely on example demonstration. 

\noindent
\textbf{Truncation \cite{xia2025tokenskip}:} 
In this method, the CoT generated by the LLM is forcibly truncated according to the preset compression ratio to limit the generation delay of the LLM.

\noindent
\textbf{CoD \cite{xu2025chain}:} 
This method drafts the concise intermediate reasoning of CoT through extensive manual design, capturing only the essential information, and guides the LLM to produce a succinct reasoning process through few-shot learning.

\noindent
\textbf{Tokenskip \cite{xia2025tokenskip}:} 
In this method, the LLMLingua-2 \cite{pan2024llmlingua} method with variable compression ratio is used to construct a short CoT dataset and perform supervised fine-tuning of LLM to realize the compression of CoT by LLM with different amplitudes.

\vspace{0.3cm}
\subsection{Implementation Details}\label{A.3}
We provide experimental details for all baseline methods in the main experiment here.
For the Prompt \cite{xia2025tokenskip} and CoD \cite{xu2025chain} methods, we show detailed prompts of these methods at different compression ratios in Table \ref{tab:prompt_other}. 
In addition, for the few-shot part of the CoD prompt, we used the original design and do not change it.
For Truncation \cite{xia2025tokenskip}, we forcibly truncate the full CoT generated by the LLM to the corresponding compression ratio, and then tested the performance of the LLM.
For Tokenskip \cite{xia2025tokenskip}, we employed its provided short CoT dataset to fine-tune the LLM using LoRA \cite{hu2022lora}, an efficient and widely validated method for LLM adaptation. The LoRA configuration used a rank $r=8$ and scaling parameter $\alpha=16$. Training was conducted for 3 epochs with a minibatch size of 4, using the AdamW \cite{loshchilov2017decoupled} optimizer and a learning rate of $5\times10^{-5}$.
All experimental results are the average results of 5 different random seeds on a single NVIDIA A100 GPU.

\begin{table*}[!h]
\small
\centering
\begin{tabular}{p{0.95\linewidth}l}
\toprule
\textbf{Prompt \cite{xia2025tokenskip} \quad Ratio:0.9 }\\ 
\bottomrule
\ttfamily  system\--{prompt}:  You are a helpful assistant. Think step by step, reduce your thought process by 10 percent.
\\
\toprule
\textbf{Prompt \cite{xia2025tokenskip} \quad Ratio:0.7 }\\ 
\bottomrule
\ttfamily  system\--{prompt}:  You are a helpful assistant. Think step by step, reduce your thought process by 30 percent.
\\ 
\toprule
\textbf{Prompt \cite{xia2025tokenskip} \quad Ratio:0.5}\\ 
\bottomrule
\ttfamily  system\--{prompt}:  You are a helpful assistant. Think step by step, reduce your thought process by 50 percent.\\ 
\toprule
\textbf{CoD \cite{xu2025chain} \quad Ratio:0.9}\\ 
\bottomrule
\ttfamily  system\--{prompt}: You are a helpful assistant. Think step by step, but only keep a minor draft for each thinking step, with 60 words at most. Return the answer at the end of the response after a separator. Put only the final answer inside \textbackslash boxed\{...\}. format: Q:{question} A: {answer} few\--{shot}: <demostration>.
\\
\toprule
\textbf{CoD \cite{xu2025chain} \quad Ratio:0.7}\\ 
\bottomrule
\ttfamily  system\--{prompt}: You are a helpful assistant. Think step by step, but only keep a minor draft for each thinking step, with 30 words at most. Return the answer at the end of the response after a separator. Put only the final answer inside \textbackslash boxed\{...\}. format: Q:{question} A: {answer} few\--{shot}: <demostration>.
\\
\toprule
\textbf{CoD \cite{xu2025chain} \quad Ratio:0.5}\\ 
\bottomrule
\ttfamily  system\--{prompt}: You are a helpful assistant. Think step by step, but only keep a minor draft for each thinking step, with 5 words at most. Return the answer at the end of the response after a separator. Put only the final answer inside \textbackslash boxed\{...\}. format: Q:{question} A: {answer} few\--{shot}: <demostration>.
\\
\bottomrule
\vspace{-1em}
\end{tabular}
\vspace{-0.5cm}
\caption{Prompt details of the main experiment in \secref{main}.}
\label{tab:prompt_other}
\end{table*}

\vspace{0.3cm}
\subsection{Training Details}\label{A.4}
In this subsection, we provide additional details for our method. 
To construct the training corpus for both the compressor and executor, we use Llama-3.1-8B-Instruct and Qwen2.5-7B-Instruct to directly generate the original CoTs by performing reasoning on the corresponding training datasets, without the need for an additional short CoT dataset. 
%For evaluation, we follow the experimental setup of \cite{xia2025tokenskip} and use the corresponding test sets to validate the final performance of each method. 
After preparing the training data, we train our compressor and executor modules in two separate stages.
In the Contextual CoT Generation stage, 
we use standard LoRA \cite{hu2022lora} to train a compressor with the Qwen2.5-1.5B as the backbone, setting the rank $r$ to 8 and the scaling parameter $\alpha$ to 16.
We adopt the AdamW to optimize both the compressor in 3 epochs of training, with a learning rate of $8\times10^{-5}$ and a minibatch size of 16.
In the Contextual CoT Utilization stage, the projector consists of two fully connected layers, $\omega_{p}^{1} \in \mathbb{R}^{|H_{c}|\times |H_{middel}|}$ and $\omega_{p}^{2} \in \mathbb{R}^{|H_{middel}|\times |H_{e}|}$, where $|H_{c}|$ and $|H_{e}|$ are the dimensions of the hidden layer of the compressor and executor respectively, and the size of the middle layer \(|H_{middel}|\) is uniformly set to 2048 in our experiment. 
We also use standard LoRA \cite{hu2022lora} to add learnable parameters to the executor, with the rank \(r\) set to 16 and the scaling parameter \(\alpha\) set to 32. 
We adopt AdamW to optimize both the executor and the projector in 3 epochs of training, with a learning rate of \(3\times10^{-5}\) and a minibatch size of 2.
During the inference phase, the max length of tokens is set to 512 for GSM8K and 1024 for MATH, and to preserve the diversity of the executor output, we set the Temperature to 0.2 and Top - p to 0.9. 
Regarding the experimental setup, all experimental results are the average results of 5 different random seeds, and the experiments are conducted on a single NVIDIA A100 GPU. 
Moreover, in Table \ref{tab:prompt_main}, we show the prompts \(z_{c}\) and \(z_{e}\) used by the compressor and executor respectively.

\begin{table}[!h]
\normalsize
\centering
\begin{tabular}{p{0.95\linewidth}l}
\toprule
\textbf{Compressor System Prompt} $z_{c}$\\ 
\bottomrule
\ttfamily  prompt: You are a helpful assistant. Analyze the question step-by-step, and and output the complete analysis process after the question.
\\
\toprule
\textbf{Executor System Prompt} $z_{e}$\\ 
\bottomrule
\ttfamily  prompt: You are a helpful assistant. Analyze the question step-by-step, and ensure your solution is highlighted within \textbackslash boxed\{...\}.
\\ 
\bottomrule
\end{tabular}
\caption{Prompt details of the UCoT.}
\label{tab:prompt_main}
\end{table}

\vspace{-0.6cm}
\subsection{Training and Evaluation Protocol}\label{A.5}
Since AIME, ASDiv, GPQA, and HumanEval do not have officially compiled standardized training datasets, their core function is positioned as model evaluation tools, we train the model solely on GSM8K and directly evaluate it on these four benchmarks without any domain-specific fine-tuning. This protocol naturally reflects a zero-shot generalization scenario and allows us to assess the method’s effectiveness across diverse tasks and domains.
During inference, generation parameters are adjusted based on task complexity. We set the maximum output length to 512 for ASDiv and 20,480 for AIME, GPQA, and HumanEval. A temperature of 0.8 and top-p of 0.9 is used across all tasks, and all test results are based on pass@1, meaning that each sample is tested only once, and if the sample passes the test, the problem is considered solve.

\subsection{Optimization Process}\label{A.6}
In the UCoT architecture, we meticulously design two crucial training stages: Contextual CoT Generation and Contextual CoT Utilization, as comprehensively illustrated in Algorithms \ref{algorithm:c} and \ref{algorithm:e}, respectively. In addition, Algorithm \ref{algorithm:i} illustrates the reasoning process of UCoT.

\begin{algorithm}
\caption{Contextual CoT Generation}
\renewcommand{\baselinestretch}{1.15}\selectfont
\label{algorithm:c}
\begin{algorithmic}[1]
\State \textbf{Input:} compressor $\mathcal{M}_{c}$ with learnable parameters $\theta_{c}$, training dataset $\mathcal{D}=\{(Q_{n},C_{n}, A_{n})\}_{n=0}^{N-1}$, learning rate $\eta$, system prompt $z_{c}$, a sequence of placeholders $s_{spec}$

\State Initialize learnable parameters $\theta_{c}$ with random values
\While{not converged}
    \For{each sample $(Q_{n},C_{n}, A_{n})$ in $\mathcal{D}$}
        \State Construct input sequence by Eq. \eqref{eq:input_form} 
        \State Get the soft tokens by Eq. \eqref{eq:ut_extract} 
        \State Compute loss $\mathcal{L}_{c}$ by Eq. \eqref{eq:loss_c}
        \State Compute gradients $\nabla \theta_{c} = \frac{\partial \mathcal{L}_{c}}{\partial \theta_{c}}$
        \State Update parameters $\theta_{c} \leftarrow \theta_{c} - \eta \nabla \theta_{c}$
    \EndFor
\EndWhile
\State \textbf{Output:} compressor $\mathcal{M}_{c}$ with the trained parameters $\theta_{c}$
\end{algorithmic}
\end{algorithm}

\begin{algorithm}
\caption{Contextual CoT Utilization}
\label{algorithm:e}
\begin{algorithmic}[1]
\State \textbf{Input:} projector $\mathcal{M}_{p}$, compressor $\mathcal{M}_{c}$ with the trained parameters $\theta_{c}$, executor $\mathcal{M}_{e}$ with learnable parameters $\theta_{e}$, training dataset $\mathcal{D}=\{(Q_{n},C_{n}, A_{n})\}_{n=0}^{N-1}$, learning rate $\eta$, system prompt $z_{c}$ and $z_{e}$, a sequence of placeholders $s_{spec}$

\State Initialize learnable parameters $\theta_{e}$ and $\mathcal{M}_{p}$ with random values
\While{not converged}
    \For{each sample $(q_{n},c_{n}, a_{n})$ in $\mathcal{D}$}
        \State Construct input sequence by Eq. \eqref{eq:input_form} 
        \State Get the soft tokens by Eq. \eqref{eq:ut_extract}
        \State Get the output $\bar{C}_{n}$ by Eq. \eqref{10}
        \State Compute loss $\mathcal{L}_{\text{sem}}$ by Eq. \eqref{11}
        \State Compute reward factor  $\mathcal{R}$ by Eq. \eqref{12}
        \State Compute loss $\mathcal{L}_{e} = \mathcal{L}_{\text{sem}} \cdot \mathcal{R}$.
        \State Compute gradients $\nabla \theta_{e} = \frac{\partial \mathcal{L}_{e}}{\partial \theta_{e}}$ and \\ \quad  \quad \quad $\nabla \mathcal{M}_{p} = \frac{\partial \mathcal{L}_{e}}{\partial \mathcal{M}_{p}}$
        \State Update parameters $\theta_{e} \leftarrow \theta_{e} - \eta \nabla \theta_{e}$ \\ \quad  \quad \quad$\mathcal{M}_{p} \leftarrow \mathcal{M}_{p} - \eta \nabla \mathcal{M}_{p}$
    \EndFor
\EndWhile
\State \textbf{Output:} executor $\mathcal{M}_{e}$ with the trained parameters $\theta_{e}$ and $\mathcal{M}_{p}$
\end{algorithmic}
\end{algorithm}

\vspace{-0.5cm}
\begin{algorithm}[H]
\caption{Efficient Post-Reasoning}
\renewcommand{\baselinestretch}{1.15}\selectfont
\label{algorithm:i}
\begin{algorithmic}[1]
\State \textbf{Input:} trained projector $\mathcal{M}_{p}$, compressor $\mathcal{M}_{c}$ with the trained parameters $\theta_{c}$, executor $\mathcal{M}_{e}$ with the trained parameters $\theta_{e}$, testing dataset $\mathcal{D}=\{(Q_{n})\}_{n=0}^{N-1}$, system prompt $z_{c}$ and $z_{e}$, placeholders $s_{spec}$
    \For{each question $q_{n}$ in $\mathcal{D}$}
        \State Construct input sequence $ (z_{c} \oplus q_{n} \oplus s_{spec} )$ 
        \State Get the soft tokens by Eq. \eqref{eq:ut_extract}
        \State Get the output by Eq. \eqref{14}
    \EndFor
\State \textbf{Output:} the answer for every question
\end{algorithmic}
\end{algorithm}

% We set the model's maximum output length to 512 for ASDiv dataset. For the more complex dataset,  AIME2024, GPQA, and HumanEval, the model's maximum output length is configured to 20480. All models use a temperature of 0.8, and we report the model's output results from a single sampling process. Specifically, for each sample in the dataset, only one random selection and model inference are performed.
\vspace{1cm}
\section{Problem Formulation}\label{B}
This section briefly introduces definitions of Chain-of-Thought Compression.

%\fakeparagraph{Chain-of-Thought Reasoning.} Given a question $Q= \{ q_{1}, q_{2}, ...,q_{\left | Q \right |}  \} $ consisting of $\left | Q \right |$ tokens, the chain-of-thought reasoning can be divided into two stages: the question reasoning stage and the question answering stage. In question reasoning stage, the model $\text{LLM}(\cdot)$ generates a step-by-step solution path $C= \{ c_{1}, c_{2}, ...,c_{\left | C \right |}  \}$ based on the question. This process can be described as:
%\begin{equation}\label{1}
%c_{i+1} = \text{LLM}(C_{\le i}; Q),
%\end{equation}
%where $C_{\le i}= \{ c_{1}, c_{2}, ...,c_{\left | i \right |}  \}$ denote the $i$ tokens in $C$ generated previously. 
%Subsequently, in question answering stage, the model leverages the question $Q$ and the previously generated solution path $R$ to solve the problem and output the final answer $A= \{ a_{1}, a_{2}, ...,a_{\left | A \right |}  \}$:
%\begin{equation}\label{2}
%a_{j+1} = \text{LLM}(A_{\le j}; Q; C),
%\end{equation}
%where $A_{\le j}= \{ a_{1}, a_{2}, ...,a_{\left | j \right |}  \}$ denote the $j$ tokens in $A$ generated previously. Previous works \cite{lee2025well,wang2024reasoning} have shown that the goal of chain-of-thought reasoning is to clarify the key elements and reasoning paths of the problem, which can effectively improve the reasoning performance of LLMs. However, overly long CoTs can lead to a significant increase in reasoning latency for LLMs.

\fakeparagraph{Chain-of-Thought Compression.} The chain-of-thought compression addresses the challenge posed by long reasoning paths in chain-of-thought reasoning by reducing the length of the reasoning path $R$ while maintaining or even enhancing the reasoning performance.
Specifically, given a question $Q=\{ q_{1}, q_{2}, ...,q_{\left | Q \right |}  \} $ and compression ratio $\alpha \in (0,1)$, chain-of-thought compression uses additional training parameters or prompt $\theta$ to guide model $\text{LLM}(\cdot)$ in generating a shorter reasoning path $\bar{C} = \{ \bar{c}_1, \bar{c}_2, \dots, \bar{c}_{|\bar{C}|} \}$: 
\begin{equation}\label{3} 
\bar{c}_{i+1} = \text{LLM}(\bar{C}_{\le i}; Q; \theta),
\end{equation}
while maintaining performance comparable to that of the original reasoning path $C$:
\begin{equation}\label{4} 
\begin{gathered}
\text{LLM}(Q; \bar{C};\theta)\approx \text{LLM}(Q; C),\\
|\bar{C}|/|C| =\alpha.
\end{gathered}
\end{equation}
Therefore, for a given dataset $\mathcal{D}=\{Q_{n},A_{n}\}_{n=1}^{N}$, consisting of $N$ question-answer pairs, the optimization objective of CoT compression with compression ratio $\alpha$ can be expressed as:
\begin{equation}\label{5}
\begin{gathered}
\theta^{*} = \arg\min_{\theta}\mathbb{E}_{\mathcal{D}} \left [ \mathbb{D}(\text{LLM}(Q; \bar{C};\theta), \text{LLM}(Q; C)) \right ] \\
    \mathrm{s.t.}\;|\bar{C}|/|C|=\alpha,
\end{gathered}
\end{equation}
where $\mathbb{D}(.)$ indicates the performance difference function of LLM before and after cot compression. In a nutshell, Eq.\ref{5} searches for parameters or prompt that minimize the difference in LLM inference performance at compression ratio $\alpha$.

\section{Additional Evaluation and Discussion on UCoT Performance}\label{C}

\subsection{Analysis of Compressor Type}\label{C.1}
We thoroughly explore the feasibility of applying different types of language models as compressors to the UCoT architecture. 
As shown in \tabref{tab:Compressor_Type}, we present the experimental results of using the Llama-3.2-1B-Instrust model as a compressor to perform the CoT compression task. 
The experimental result shows that under this setting, UCoT can still effectively reduce the length of the CoT output by the executor while maintaining the inference accuracy of the executor. 
When compared with the results of our main experiment (\tabref{table:main}), UCoT still demonstrates significant competitive advantages over other comparative methods. 
The above results fully verify the adaptability of the compressor component in the UCoT architecture to different types of language models, indicating that this design has good generality and extensibility.

\vspace{0.15cm}
\subsection{Analysis of Executor Scale}\label{C.2}
In the experiment of this section, we systematically explored the performance of the UCoT method when applied to executors of different scales. 
As shown in \tabref{tab:Executor_Scale}, we select Qwen2.5 models with four different parameter magnitudes of 1.5B, 3B, 7B, and 14B as executors and evaluated their performance on the GSM8K dataset. The experimental results indicate that the UCoT method can efficiently complete the compression of CoT for executors of different scales. 
Specifically, as the scale of the executor increases, the executor can better maintain its performance under the same compression ratio condition. 
Taking the Qwen2.5-14B model as an example, when the UCoT method reduces the token output by 48\%, the accuracy of the model only decreases by 1.79\%. 
This phenomenon fully demonstrates that a larger-scale Large Language Model (LLM) can more effectively extract and utilize the key reasoning information in soft tokens, thus still maintaining a high performance level after compression.

%%%%%%%%%%%%% Table-Llama3.2-1B %%%%%%%%%%%%%
\begin{table*}[!h]
\centering
\renewcommand\arraystretch{0.7}
\resizebox{\textwidth}{!}{
\begin{tabular}{lc|lccc|lccc}
\toprule \midrule
        &  \multicolumn{1}{c}{}  &  \multicolumn{4}{c}{\textbf{GSM8K}} & \multicolumn{4}{c}{\textbf{MATH-500}}  \\ \cmidrule(lr){3-6} \cmidrule(lr){7-10}
\textbf{Methods} & \multicolumn{1}{c}{\textbf{Ratio}} & Acc.\ (\%)\ $\uparrow$ & \ Tokens\ $\downarrow$ & \multicolumn{1}{c}{ Latency\ (s)\ $\downarrow$ }& ActRatio\ $\downarrow$ & Acc.\ (\%)\ $\uparrow$ & \ Tokens\ $\downarrow$ & \multicolumn{1}{c}{ Latency\ (s)\ $\downarrow$ }& ActRatio\ $\downarrow$
\\ \specialrule{0.05em}{0.3em}{0.1em}
\rowcolor[gray]{0.9}
\multicolumn{10}{c}{\textbf{\textit{Qwen2.5-7B Series}}}
\\ \specialrule{0.05em}{0.1em}{0.3em}
\multirow{1}{*}{\makecell[c]{Original}} 
&-       & 92.17$_{\textcolor{Maroon}{\text{0.00}\downarrow}}$               
         & 298.63              
         & 3.83$_{\textcolor{ForestGreen}{\text{1.00}\times}}$                       
         & -
         & 61.60$_{\textcolor{Maroon}{\text{0.00}\downarrow}}$               
         & 571.64              
         & 6.35$_{\textcolor{ForestGreen}{\text{1.00}\times}}$                       
         & -
\\ \midrule

\multirow{3}{*}{\makecell[c]{UCoT}}
&0.9     & 90.58$_{\textcolor{Maroon}{\text{1.59}\downarrow}}$
         & 271.75                            
         & 3.52$_{\textcolor{ForestGreen}{\text{1.09}\times}}$  
         & 0.91
         & 59.80$_{\textcolor{Maroon}{\text{1.80}\downarrow}}$
         & 485.91                            
         & 5.53$_{\textcolor{ForestGreen}{\text{1.15}\times}}$  
         & 0.85          \\ 
&0.7     & 87.62$_{\textcolor{Maroon}{\text{4.55}\downarrow}}$
         & 206.05                            
         & 2.71$_{\textcolor{ForestGreen}{\text{1.41}\times}}$  
         & 0.69          
         & 58.40$_{\textcolor{Maroon}{\text{3.20}\downarrow}}$
         & 394.44                            
         & 4.41$_{\textcolor{ForestGreen}{\text{1.44}\times}}$  
         & 0.69
         \\ 
&0.5     & 86.55$_{\textcolor{Maroon}{\text{5.62}\downarrow}}$    
         & 155.29         
         & 2.04$_{\textcolor{ForestGreen}{\text{1.88}\times}}$ 
         & 0.52
         & 53.20$_{\textcolor{Maroon}{\text{8.40}\downarrow}}$    
         & 285.82         
         & 3.18$_{\textcolor{ForestGreen}{\text{2.00}\times}}$ 
         & 0.50
\\ \specialrule{0.05em}{0.3em}{0.1em}

\rowcolor[gray]{0.9}
\multicolumn{10}{c}{\textbf{\textit{Llama3.1-8B Series}}}
\\ \specialrule{0.05em}{0.1em}{0.3em}

\multirow{1}{*}{\makecell[c]{Original}}
&-        & 87.26$_{\textcolor{Maroon}{\text{0.00}\downarrow}}$
          & 212.13
          & 2.44$_{\textcolor{ForestGreen}{\text{1.00}\times}}$
          & -
          & 42.70$_{\textcolor{Maroon}{\text{0.00}\downarrow}}$
          & 574.28
          & 6.88$_{\textcolor{ForestGreen}{\text{1.00}\times}}$
          & -
\\ \midrule

\multirow{3}{*}{\makecell[c]{UCoT}}
&0.9     & 86.32$_{\textcolor{Maroon}{\text{0.94}\downarrow}}$ 
         & 186.65                        
         & 2.21$_{\textcolor{ForestGreen}{\text{1.10}\times}}$
         & 0.88 
         & 40.30$_{\textcolor{Maroon}{\text{2.40}\downarrow}}$ 
         & 493.86                        
         & 6.06$_{\textcolor{ForestGreen}{\text{1.14}\times}}$
         & 0.86
         \\ 
&0.7     & 83.97$_{\textcolor{Maroon}{\text{3.29}\downarrow}}$ 
         & 154.58                         
         & 1.82$_{\textcolor{ForestGreen}{\text{1.34}\times}}$
         & 0.73 
         & 39.20$_{\textcolor{Maroon}{\text{3.50}\downarrow}}$ 
         & 384.77                         
         & 4.73$_{\textcolor{ForestGreen}{\text{1.45}\times}}$
         & 0.67
         \\ 
&0.5     & 83.16$_{\textcolor{Maroon}{\text{4.10}\downarrow}}$
         & 114.52                
         & 1.42$_{\textcolor{ForestGreen}{\text{1.72}\times}}$
         & 0.54
         & 37.30$_{\textcolor{Maroon}{\text{5.40}\downarrow}}$
         & 287.16                
         & 3.74$_{\textcolor{ForestGreen}{\text{1.84}\times}}$
         & 0.50
\\ \midrule
\bottomrule
\end{tabular}}
\caption{Performance comparison of compressor type.}

\label{tab:Compressor_Type}
\end{table*}
%%%%%%%%%%% Table-Llama3.2-1B %%%%%%%%%%%%%%%%%%%

%%%%%%%%%%%%% Table-Model-Size %%%%%%%%%%%%%
\begin{table*}[!h]
\centering
\renewcommand\arraystretch{0.7}
\resizebox{\textwidth}{!}{
\begin{tabular}{lc|lccc|lccc}
\toprule \midrule
        &  \multicolumn{1}{c}{}  &  \multicolumn{8}{c}{\textbf{GSM8K}} \\ \cmidrule(lr){3-10} 
\textbf{Methods} & \multicolumn{1}{c}{\textbf{Ratio}}       & Acc.\ (\%)\ $\uparrow$ & \ Tokens\ $\downarrow$ & \multicolumn{1}{c}{ Latency\ (s)\ $\downarrow$ }& ActRatio\ $\downarrow$ & Acc.\ (\%)\ $\uparrow$ & \ Tokens\ $\downarrow$ & Latency\ (s)\ $\downarrow$ & ActRatio\ $\downarrow$ 

\\ \specialrule{0.05em}{0.3em}{0.1em}
\rowcolor[gray]{0.9}
\multicolumn{2}{c}{} &
\multicolumn{4}{|c|}{\textbf{\textit{Qwen2.5-1.5B Series}}} &
\multicolumn{4}{c}{\textbf{\textit{Qwen2.5-3B Series}}}
\\ \specialrule{0.05em}{0.1em}{0.3em}
\multirow{1}{*}{\makecell[c]{Original}} 
&-       & 70.96$_{\textcolor{Maroon}{\text{0.00}\downarrow}}$               
         & 317.14              
         & 3.67$_{\textcolor{ForestGreen}{\text{1.00}\times}}$                       
         & -            
         & 83.62$_{\textcolor{Maroon}{\text{0.00}\downarrow}}$     
         & 316.83 
         & 3.71$_{\textcolor{ForestGreen}{\text{1.00}\times}}$                  
         & -
\\ \midrule

\multirow{3}{*}{\makecell[c]{UCoT}}
&0.9     & 68.77$_{\textcolor{Maroon}{\text{2.19}\downarrow}}$
         & 291.77                            
         & 3.41$_{\textcolor{ForestGreen}{\text{1.08}\times}}$
         & 0.92  
         & 83.45$_{\textcolor{Maroon}{\text{0.17}\downarrow}}$ 
         & 275.64               
         & 3.40$_{\textcolor{ForestGreen}{\text{1.09}\times}}$
         & 0.87          \\ 
&0.7     & 64.15$_{\textcolor{Maroon}{\text{6.81}\downarrow}}$
         & 228.34                            
         & 2.75$_{\textcolor{ForestGreen}{\text{1.33}\times}}$  
         & 0.72  
         & 81.52$_{\textcolor{Maroon}{\text{2.10}\downarrow}}$ 
         & 224.95               
         & 2.82$_{\textcolor{ForestGreen}{\text{1.32}\times}}$
         & 0.71          \\ 
&0.5     & 61.58$_{\textcolor{Maroon}{\text{9.38}\downarrow}}$    
         & 155.40         
         & 1.83$_{\textcolor{ForestGreen}{\text{2.01}\times}}$ 
         & 0.49  
         & 80.67$_{\textcolor{Maroon}{\text{2.95}\downarrow}}$    
         & 164.75                
         & 2.07$_{\textcolor{ForestGreen}{\text{1.79}\times}}$
         & 0.52
\\ \specialrule{0.05em}{0.3em}{0.1em}

\rowcolor[gray]{0.9}
\multicolumn{2}{c}{} &
\multicolumn{4}{|c|}{\textbf{\textit{Qwen2.5-7B Series}}} &
\multicolumn{4}{c}{\textbf{\textit{Qwen2.5-14B Series}}}
\\ \specialrule{0.05em}{0.1em}{0.3em}

\multirow{1}{*}{\makecell[c]{Original}}
&-        & 92.17$_{\textcolor{Maroon}{\text{0.00}\downarrow}}$
          & 298.63
          & 4.28$_{\textcolor{ForestGreen}{\text{1.00}\times}}$
          & -                            
          & 93.12$_{\textcolor{Maroon}{\text{0.00}\downarrow}}$     
          & 314.37
          & 4.57$_{\textcolor{ForestGreen}{\text{1.00}\times}}$
          & -
\\ \midrule

\multirow{3}{*}{\makecell[c]{UCoT}}
&0.9     & 91.72$_{\textcolor{Maroon}{\text{0.45}\downarrow}}$ 
         & 244.88                 
         & 3.72$_{\textcolor{ForestGreen}{\text{1.15}\times}}$
         & 0.85 
         & 92.86$_{\textcolor{Maroon}{\text{0.26}\downarrow}}$ 
         & 279.79            
         & 3.63$_{\textcolor{ForestGreen}{\text{1.26}\times}}$      
         & 0.89          \\ 
&0.7     & 87.98$_{\textcolor{Maroon}{\text{4.19}\downarrow}}$ 
         & 194.63                         
         & 2.83$_{\textcolor{ForestGreen}{\text{1.51}\times}}$
         & 0.65 
         & 91.72$_{\textcolor{Maroon}{\text{1.40}\downarrow}}$ 
         & 213.77            
         & 2.79$_{\textcolor{ForestGreen}{\text{1.64}\times}}$ 
         & 0.68          \\ 
&0.5     & 87.55$_{\textcolor{Maroon}{\text{4.62}\downarrow}}$  
         & 140.36                
         & 2.25$_{\textcolor{ForestGreen}{\text{1.90}\times}}$
         & 0.48 
         & 91.33$_{\textcolor{Maroon}{\text{1.79}\downarrow}}$ 
         & 169.76 
         & 2.36$_{\textcolor{ForestGreen}{\text{1.94}\times}}$
         & 0.54
\\ \midrule
\bottomrule
\end{tabular}}
\caption{Performance comparison of executor size.}

\label{tab:Executor_Scale}
\end{table*}
%%%%%%%%%%% Table-Model-size %%%%%%%%%%%%%%%%%%%

%%%%%%%%%%%%% Table-Instruction-Robustness-0.7-1 %%%%%%%%%%%%%
\begin{table*}[h!]
\centering
\renewcommand\arraystretch{0.7}
\resizebox{\textwidth}{!}{
\begin{tabular}{cc|lccc|lccc}
\toprule \midrule
        &  \multicolumn{1}{c}{}  &  \multicolumn{4}{c}{\textbf{GSM8K}}   & \multicolumn{4}{c}{\textbf{MATH-500}} \\ \cmidrule(lr){3-6} \cmidrule(lr){7-10}
\textbf{Methods} & \multicolumn{1}{c}{\textbf{Types}}       & Acc.\ (\%)\ $\uparrow$ & \ Tokens\ $\downarrow$ & \multicolumn{1}{c}{ Latency\ (s)\ $\downarrow$ }& ActRatio\ $\downarrow$ & Acc.\ (\%)\ $\uparrow$ & \ Tokens\ $\downarrow$ & Latency\ (s)\ $\downarrow$ & ActRatio\ $\downarrow$ 

\\ \specialrule{0.05em}{0.3em}{0.1em}
\rowcolor[gray]{0.9}
\multicolumn{10}{c}{\textbf{\textit{Qwen2.5-7B Series}}}
\\ \specialrule{0.05em}{0.1em}{0.3em}
\multirow{1}{*}{\makecell[c]{Original}} 
&-       & 92.17$_{\textcolor{Maroon}{\text{0.00}\downarrow}}$               
         & 298.63              
         & 3.83$_{\textcolor{ForestGreen}{\text{1.00}\times}}$                       
         & -
         & 61.60$_{\textcolor{Maroon}{\text{0.00}\downarrow}}$               
         & 571.64              
         & 6.35$_{\textcolor{ForestGreen}{\text{1.00}\times}}$                       
         & -
\\ \midrule

\multirow{3}{*}{\makecell[c]{Tokenskip \\ \cite{xia2025tokenskip}}}
&I      & 85.69$_{\textcolor{Maroon}{\text{6.48}\downarrow}}$ 
         & 223.97
         & 2.89$_{\textcolor{ForestGreen}{\text{1.33}\times}}$   
         & 0.75               
         & 56.60$_{\textcolor{Maroon}{\text{5.00}\downarrow}}$     
         & 417.29 
         & 4.72$_{\textcolor{ForestGreen}{\text{1.35}\times}}$                  
         & 0.73            \\
&P      & 78.93$_{\textcolor{Maroon}{\text{13.24}\downarrow}}$               
         & 215.01              
         & 2.79$_{\textcolor{ForestGreen}{\text{1.37}\times}}$   
         & 0.72               
         & 52.70$_{\textcolor{Maroon}{\text{8.90}\downarrow}}$
         & 402.76 
         & 4.53$_{\textcolor{ForestGreen}{\text{1.40}\times}}$                  
         & 0.70           \\
&C      & 80.11$_{\textcolor{Maroon}{\text{12.06}\downarrow}}$               
         & 218.00              
         & 2.82$_{\textcolor{ForestGreen}{\text{1.36}\times}}$                       
         & 0.73               
         & 51.30$_{\textcolor{Maroon}{\text{10.30}\downarrow}}$     
         & 394.43
         & 4.50$_{\textcolor{ForestGreen}{\text{1.41}\times}}$    
         & 0.69
\\ \midrule

\multirow{3}{*}{\makecell[c]{UCoT}}
&I      & 87.98$_{\textcolor{Maroon}{\text{4.19}\downarrow}}$
         & 194.11                            
         & 2.65$_{\textcolor{ForestGreen}{\text{1.45}\times}}$
         & 0.65
         & 58.80$_{\textcolor{Maroon}{\text{2.80}\downarrow}}$
         & 388.71              
         & 4.46$_{\textcolor{ForestGreen}{\text{1.42}\times}}$
         & 0.68         \\ 
&P      & 84.62$_{\textcolor{Maroon}{\text{7.55}\downarrow}}$
         & 200.08                            
         & 2.69$_{\textcolor{ForestGreen}{\text{1.42}\times}}$  
         & 0.67  
         & 56.30$_{\textcolor{Maroon}{\text{5.30}\downarrow}}$ 
         & 405.86               
         & 4.53$_{\textcolor{ForestGreen}{\text{1.40}\times}}$
         & 0.71          \\ 
&C      & 85.39$_{\textcolor{Maroon}{\text{6.78}\downarrow}}$    
         & 203.17         
         & 2.71$_{\textcolor{ForestGreen}{\text{1.41}\times}}$ 
         & 0.68  
         & 56.70$_{\textcolor{Maroon}{\text{4.90}\downarrow}}$    
         & 402.88                
         & 4.55$_{\textcolor{ForestGreen}{\text{1.40}\times}}$
         & 0.70
\\ \specialrule{0.05em}{0.3em}{0.1em}

\bottomrule
\end{tabular}}
\caption{Performance comparison of prompt robustness in Qwen2.5-7B series.}

\label{tab:robust_qwen}
\end{table*}
%%%%%%%%%%% Table-Instruction-Robustness-0.7-1 %%%%%%%%%%%%%%%%%%%

%%%%%%%%%%%%% Table-Instruction-Robustness-0.7 %%%%%%%%%%%%%
\begin{table*}[!h]
\centering
\renewcommand\arraystretch{0.7}
\resizebox{\textwidth}{!}{
\begin{tabular}{cc|lccc|lccc}
\toprule \midrule
        &  \multicolumn{1}{c}{}  &  \multicolumn{4}{c}{\textbf{GSM8K}}   & \multicolumn{4}{c}{\textbf{MATH-500}} \\ \cmidrule(lr){3-6} \cmidrule(lr){7-10}
\textbf{Methods} & \multicolumn{1}{c}{\textbf{Types}}       & Acc.\ (\%)\ $\uparrow$ & \ Tokens\ $\downarrow$ & \multicolumn{1}{c}{ Latency\ (s)\ $\downarrow$ }& ActRatio\ $\downarrow$ & Acc.\ (\%)\ $\uparrow$ & \ Tokens\ $\downarrow$ & Latency\ (s)\ $\downarrow$ & ActRatio\ $\downarrow$ 

\\ \specialrule{0.05em}{0.3em}{0.1em}
\rowcolor[gray]{0.9}
\multicolumn{10}{c}{\textbf{\textit{Llama3.1-8B Series}}}
\\ \specialrule{0.05em}{0.1em}{0.3em}

\multirow{1}{*}{\makecell[c]{Original}}
&-        & 87.26$_{\textcolor{Maroon}{\text{0.00}\downarrow}}$
          & 212.13
          & 2.44$_{\textcolor{ForestGreen}{\text{1.00}\times}}$
          & -
          & 42.70$_{\textcolor{Maroon}{\text{0.00}\downarrow}}$
          & 574.28
          & 6.88$_{\textcolor{ForestGreen}{\text{1.00}\times}}$
          & -
\\ \midrule

\multirow{3}{*}{\makecell[c]{Tokenskip \\ \cite{xia2025tokenskip}}}
&I       & 82.57$_{\textcolor{Maroon}{\text{4.69}\downarrow}}$
          & 161.21
          & 1.86$_{\textcolor{ForestGreen}{\text{1.31}\times}}$
          & 0.76                            
          & 37.90$_{\textcolor{Maroon}{\text{4.80}\downarrow}}$     
          & 413.48
          & 4.95$_{\textcolor{ForestGreen}{\text{1.39}\times}}$
          & 0.72       \\
&P       & 78.32$_{\textcolor{Maroon}{\text{8.94}\downarrow}}$
          & 150.62
          & 1.79$_{\textcolor{ForestGreen}{\text{1.36}\times}}$
          & 0.71                    
          & 34.70$_{\textcolor{Maroon}{\text{8.00}\downarrow}}$
          & 431.02
          & 5.23$_{\textcolor{ForestGreen}{\text{1.32}\times}}$
          & 0.75       \\
&C       & 79.68$_{\textcolor{Maroon}{\text{7.58}\downarrow}}$
          & 157.11
          & 1.84$_{\textcolor{ForestGreen}{\text{1.33}\times}}$
          & 0.74                            
          & 36.70$_{\textcolor{Maroon}{\text{6.00}\downarrow}}$     
          & 442.26
          & 5.32$_{\textcolor{ForestGreen}{\text{1.29}\times}}$
          & 0.77
\\ \midrule

\multirow{3}{*}{\makecell[c]{UCoT}}
&I      & 84.17$_{\textcolor{Maroon}{\text{3.09}\downarrow}}$ 
         & 140.01                 
         & 1.72$_{\textcolor{ForestGreen}{\text{1.42}\times}}$
         & 0.66
         & 39.50$_{\textcolor{Maroon}{\text{3.20}\downarrow}}$
         & 396.25           
         & 4.80$_{\textcolor{ForestGreen}{\text{1.43}\times}}$      
         & 0.69          \\ 
&P      & 82.55$_{\textcolor{Maroon}{\text{4.71}\downarrow}}$ 
         & 144.25                         
         & 1.74$_{\textcolor{ForestGreen}{\text{1.40}\times}}$
         & 0.68 
         & 37.80$_{\textcolor{Maroon}{\text{4.90}\downarrow}}$ 
         & 402.00            
         & 4.97$_{\textcolor{ForestGreen}{\text{1.38}\times}}$ 
         & 0.70          \\ 
&C      & 81.97$_{\textcolor{Maroon}{\text{5.29}\downarrow}}$  
         & 146.37
         & 1.74$_{\textcolor{ForestGreen}{\text{1.40}\times}}$
         & 0.69 
         & 39.10$_{\textcolor{Maroon}{\text{3.60}\downarrow}}$ 
         & 413.49 
         & 5.19$_{\textcolor{ForestGreen}{\text{1.33}\times}}$
         & 0.72
\\ \midrule

\bottomrule
\end{tabular}}
\caption{Performance comparison of prompt robustness in Llama3.1-8B series.}

\label{tab:robust_llama}
\end{table*}
%%%%%%%%%%% Table-Instruction-Robustness-0.7-2 %%%%%%%%%%%%%%%%%%%
\vspace{0.15cm}
\subsection{Analysis of Prompt Robustness}\label{C.3}
Currently, the research on the robustness of the CoT compression method is still limited. 
In this part, we carry out two common perturbation attacks, Paragraphing \cite{2024arXiv240211638W} and  Character-Substituted \cite{2024arXiv240211638W}, on the input prompts of Tokenskip and UCoT, aiming to observe whether they can still effectively perform the CoT compression task \footnote{Paragraphing method uses the Pegasus encoder \cite{2019arXiv191208777Z} to restructure input prompts. Character-Substituted method replaces part of the characters in the prompt according to predetermined rules. For simplicity, the initial settings of the two methods, as well as the perturbation attacks Paragraphing and Character-Substituted, are referred to and presented by I, P, and C in the relevant tables.}. 
As shown in \tabref{tab:robust_qwen} and \tabref{tab:robust_llama}, we conducted the experiment under the experimental setting with a compression ratio of $0.7$. 
The experimental results indicate that when facing Paragraphing \cite{2024arXiv240211638W} and  Character-Substituted \cite{2024arXiv240211638W} attacks, both UCoT and Tokenskip can complete the CoT compression task, but UCoT demonstrates better robustness.
This indicates that UCoT method, leveraging continuous embeddings, can more effectively mitigate the impact of perturbation attacks, which enhances its viability for real-world deployment.

%%%%%%%%%%%%% Supplementary Table1 %%%%%%%%%%%%%
\begin{table*}[!h]
\centering
\renewcommand\arraystretch{0.9}
\resizebox{\textwidth}{!}{
\begin{tabular}{cc|lccc|lccc}
\toprule \midrule
        &  \multicolumn{1}{c}{}  &  \multicolumn{4}{c}{\textbf{AIME 2024}}   & \multicolumn{4}{c}{\textbf{ASDiv}} \\ \cmidrule(lr){3-6} \cmidrule(lr){7-10}
\textbf{Methods} & \multicolumn{1}{c}{\textbf{Types}}       & Acc.\ (\%)\ $\uparrow$ & \ Tokens\ $\downarrow$ & \multicolumn{1}{c}{ Latency\ (s)\ $\downarrow$ }& ActRatio\ $\downarrow$ & Acc.\ (\%)\ $\uparrow$ & \ Tokens\ $\downarrow$ & Latency\ (s)\ $\downarrow$ & ActRatio\ $\downarrow$ 

\\ \specialrule{0.05em}{0.3em}{0.1em}
\rowcolor[gray]{0.9}
\multicolumn{10}{c}{\textbf{\textit{Qwen3-8B}}}
\\ \specialrule{0.05em}{0.1em}{0.3em}
\multirow{1}{*}{\makecell[c]{ Original}}
&-       & 66.67$_{\textcolor{Maroon}{\text{0.00}\downarrow}}$               
         & 6357.13              
         & 418.07$_{\textcolor{ForestGreen}{\text{1.00}\times}}$                       
         & -               
         & 93.74$_{\textcolor{Maroon}{\text{0.00}\downarrow}}$     
         & 475.49 
         & 4.72$_{\textcolor{ForestGreen}{\text{1.00}\times}}$                  
         & -
\\ \midrule

\multirow{3}{*}{\makecell[c]{Tokenskip \\ \cite{xia2025tokenskip}}}
&0.9     & \textcolor{red}{66.67}$_{\textcolor{Maroon}{\text{0.00}\downarrow}}$ & 5912.13                                  & 393.51$_{\textcolor{ForestGreen}{\text{1.06}\times}}$
         & 0.93   
         & 92.55$_{\textcolor{Maroon}{\text{1.19}\downarrow}}$ & 446.96         
         & 4.44$_{\textcolor{ForestGreen}{\text{1.06}\times}}$
         & 0.94                   \\ 
&0.7     & 60.00$_{\textcolor{Maroon}{\text{6.67}\downarrow}}$ & 4765.88                                  & 320.55$_{\textcolor{ForestGreen}{\text{1.30}\times}}$
         & 0.75           
         & \textcolor{red}{93.48}$_{\textcolor{Maroon}{\text{0.26}\downarrow}}$ & 361.37                         & 3.59$_{\textcolor{ForestGreen}{\text{1.31}\times}}$
         & 0.76                   \\ 
&0.5     & \textcolor{red}{56.67}$_{\textcolor{Maroon}{\text{10.0}\downarrow}}$ 
         & 4499.99                        & 307.49$_{\textcolor{ForestGreen}{\text{1.36}\times}}$
         & 0.70  
         & 91.57$_{\textcolor{Maroon}{\text{2.17}\downarrow}}$
         & 294.83                         
         & 2.93$_{\textcolor{ForestGreen}{\text{1.61}\times}}$
         & 0.62
\\ \midrule

\multirow{3}{*}{\makecell[c]{UCoT}}
&0.9     & \textcolor{red}{66.67}$_{\textcolor{Maroon}{\text{0.00}\downarrow}}$        
         & \textcolor{red}{5662.74}                
         & \textcolor{red}{382.23}$_{\textcolor{ForestGreen}{\text{1.09}\times}}$
         & \textcolor{red}{0.89}  
         & \textcolor{red}{93.53}$_{\textcolor{Maroon}{\text{0.21}\downarrow}}$ & \textcolor{red}{413.67}  
         & \textcolor{red}{4.12}$_{\textcolor{ForestGreen}{\text{1.15}\times}}$
         & \textcolor{red}{0.87}          \\ 
&0.7     & \textcolor{red}{63.33}$_{\textcolor{Maroon}{\text{3.34}\downarrow}}$
         & \textcolor{red}{4513.79}                            
         & \textcolor{red}{298.97}$_{\textcolor{ForestGreen}{\text{1.40}\times}}$  
         & \textcolor{red}{0.71}  
         & 93.32$_{\textcolor{Maroon}{\text{0.42}\downarrow}}$ & \textcolor{red}{304.31}               
         & \textcolor{red}{3.03}$_{\textcolor{ForestGreen}{\text{1.56}\times}}$
         & \textcolor{red}{0.64}          \\ 
&0.5     & \textcolor{red}{56.67}$_{\textcolor{Maroon}{\text{10.0}\downarrow}}$    
         & \textcolor{red}{4132.18}         
         & \textcolor{red}{281.55}$_{\textcolor{ForestGreen}{\text{1.48}\times}}$ 
         & \textcolor{red}{0.65}  
         & \textcolor{red}{92.77}$_{\textcolor{Maroon}{\text{0.97}\downarrow}}$    
         & \textcolor{red}{257.76}                
         & \textcolor{red}{2.57}$_{\textcolor{ForestGreen}{\text{1.84}\times}}$
         & \textcolor{red}{0.54}
\\ \specialrule{0.05em}{0.3em}{0.1em}

\rowcolor[gray]{0.9}
\multicolumn{10}{c}{\textbf{\textit{Deepseek-R1-Distill-Qwen-7B}}}
\\ \specialrule{0.05em}{0.1em}{0.3em}

\multirow{1}{*}{\makecell[c]{Original}}
&-        & 53.33$_{\textcolor{Maroon}{\text{0.00}\downarrow}}$
          & 1286.16
          & 417.46$_{\textcolor{ForestGreen}{\text{1.00}\times}}$
          & -                            
          & 95.47$_{\textcolor{Maroon}{\text{0.00}\downarrow}}$     
          & 468.94
          & 5.48$_{\textcolor{ForestGreen}{\text{1.00}\times}}$
          & -
\\ \midrule

\multirow{3}{*}{\makecell[c]{Tokenskip \\ \cite{xia2025tokenskip}}}
&0.9     & 46.67$_{\textcolor{Maroon}{\text{6.66}\downarrow}}$ & 1208.24                       & 402.41$_{\textcolor{ForestGreen}{\text{1.04}\times}}$ 
         & 0.94   
         & 95.18$_{\textcolor{Maroon}{\text{0.29}\downarrow}}$  & 431.55          
         & 5.07$_{\textcolor{ForestGreen}{\text{1.08}\times}}$
         & 0.92                   \\ 
&0.7     & 43.33$_{\textcolor{Maroon}{\text{10.0}\downarrow}}$ & 9774.88                        & 320.44$_{\textcolor{ForestGreen}{\text{1.30}\times}}$
         & 0.76           
         & 94.76$_{\textcolor{Maroon}{\text{0.71}\downarrow}}$  & 337.63               & 3.95$_{\textcolor{ForestGreen}{\text{1.39}\times}}$
         & 0.72                   \\ 
&0.5     & \textcolor{red}{43.33}$_{\textcolor{Maroon}{\text{10.0}\downarrow}}$ 
         & 9260.35                        & 286.88$_{\textcolor{ForestGreen}{\text{1.46}\times}}$
         & 0.72  
         & 94.33$_{\textcolor{Maroon}{\text{1.14}\downarrow}}$ & 301.32                & 3.52$_{\textcolor{ForestGreen}{\text{1.56}\times}}$
         & 0.64
\\ \midrule

\multirow{3}{*}{\makecell[c]{UCoT}}
&0.9     & \textcolor{red}{50.00}$_{\textcolor{Maroon}{\text{3.33}\downarrow}}$ 
         & \textcolor{red}{1093.34}
         & \textcolor{red}{342.96}$_{\textcolor{ForestGreen}{\text{1.22}\times}}$
         & \textcolor{red}{0.85}  
         & \textcolor{red}{95.43}$_{\textcolor{Maroon}{\text{0.04}\downarrow}}$  & \textcolor{red}{384.61}
         & \textcolor{red}{4.52}$_{\textcolor{ForestGreen}{\text{1.21}\times}}$
         & \textcolor{red}{0.82}          \\ 
&0.7     & \textcolor{red}{46.67}$_{\textcolor{Maroon}{\text{6.66}\downarrow}}$ 
         & \textcolor{red}{8875.30}
         & \textcolor{red}{307.05}$_{\textcolor{ForestGreen}{\text{1.36}\times}}$
         & \textcolor{red}{0.69} 
         & \textcolor{red}{96.01}$_{\textcolor{Maroon}{\text{0.54}\uparrow}}$ 
         & \textcolor{red}{300.12}            
         & \textcolor{red}{3.52}$_{\textcolor{ForestGreen}{\text{1.56}\times}}$    
         & \textcolor{red}{0.64}          \\ 
&0.5     & \textcolor{red}{43.33}$_{\textcolor{Maroon}{\text{10.0}\downarrow}}$  
         & \textcolor{red}{8231.43}                
         & \textcolor{red}{274.18}$_{\textcolor{ForestGreen}{\text{1.52}\times}}$
         & \textcolor{red}{0.64} 
         & \textcolor{red}{96.58}$_{\textcolor{Maroon}{\text{1.11}\uparrow}}$ 
         & \textcolor{red}{257.93} 
         & \textcolor{red}{3.01}$_{\textcolor{ForestGreen}{\text{1.82}\times}}$
         & \textcolor{red}{0.55}
\\ \midrule
\bottomrule
\end{tabular}}
\caption{Performance comparison of Qwen3-8B model and Deepseek-R1-Distill-Qwen-7B model on both AIME 2024 and AsDiv tasks.}

\label{table:supplementary_main_1}
\end{table*}
%%%%%%%%%%% Supplementary Table1 %%%%%%%%%%%%%%%%%%%
\subsection{Performance comparison on AIME2024 and ASDiv Datasets}\label{C.9}
This section provides additional evaluations of two difficulty levels for mathematical reasoning tasks: AIME 2024 \cite{finkelstein2024artificial} and ASDiv \cite{miao-etal-2020-diverse}. Results in Table \ref{table:supplementary_main_1} show UCoT exhibits excellent CoT compression performance, whether on the complex AIME 2024 or simpler ASDiv. For example, when UCoT is applied to the deepseek-r1-distill-qwen-7b model, the usage of inference tokens on the ASDiv dataset is reduced by 45\%, while the performance is improved by 2.25\% over the SOTA method Tokenskip.

\subsection{The Case Studies for Contextual CoT}\label{C.7}
\begin{figure*}[!h]
    \centering
    \resizebox{0.8\textwidth}{!}{%
        \includegraphics{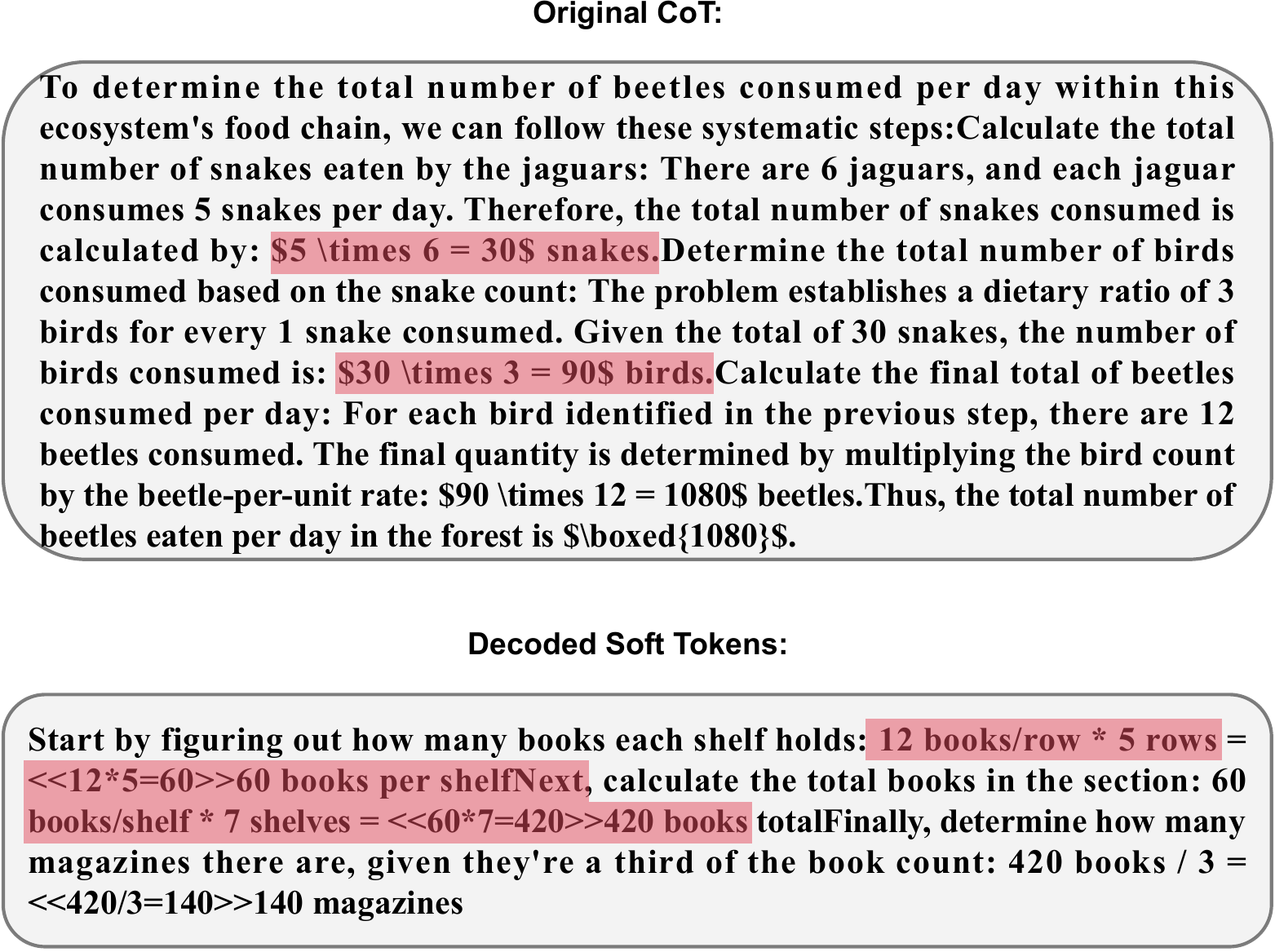} 
    }

    \caption{A case study on GSM8K dataset. Output using Qwen2.5-7B-Instruct as the executor and Qwen2.5-1.5B as the compressor. The structural information about reasoning that appear simultaneously in both the original CoT and the decoded soft tokens are highlighted in red \raisebox{0.5ex}{\colorbox{MyHighlight}{\rule{-1pt}{0pt}}}.}
    \label{fig:case_ut_1} 
\end{figure*}

\begin{figure*}[!h]
    \centering
    \resizebox{0.78\textwidth}{!}{%
        \includegraphics{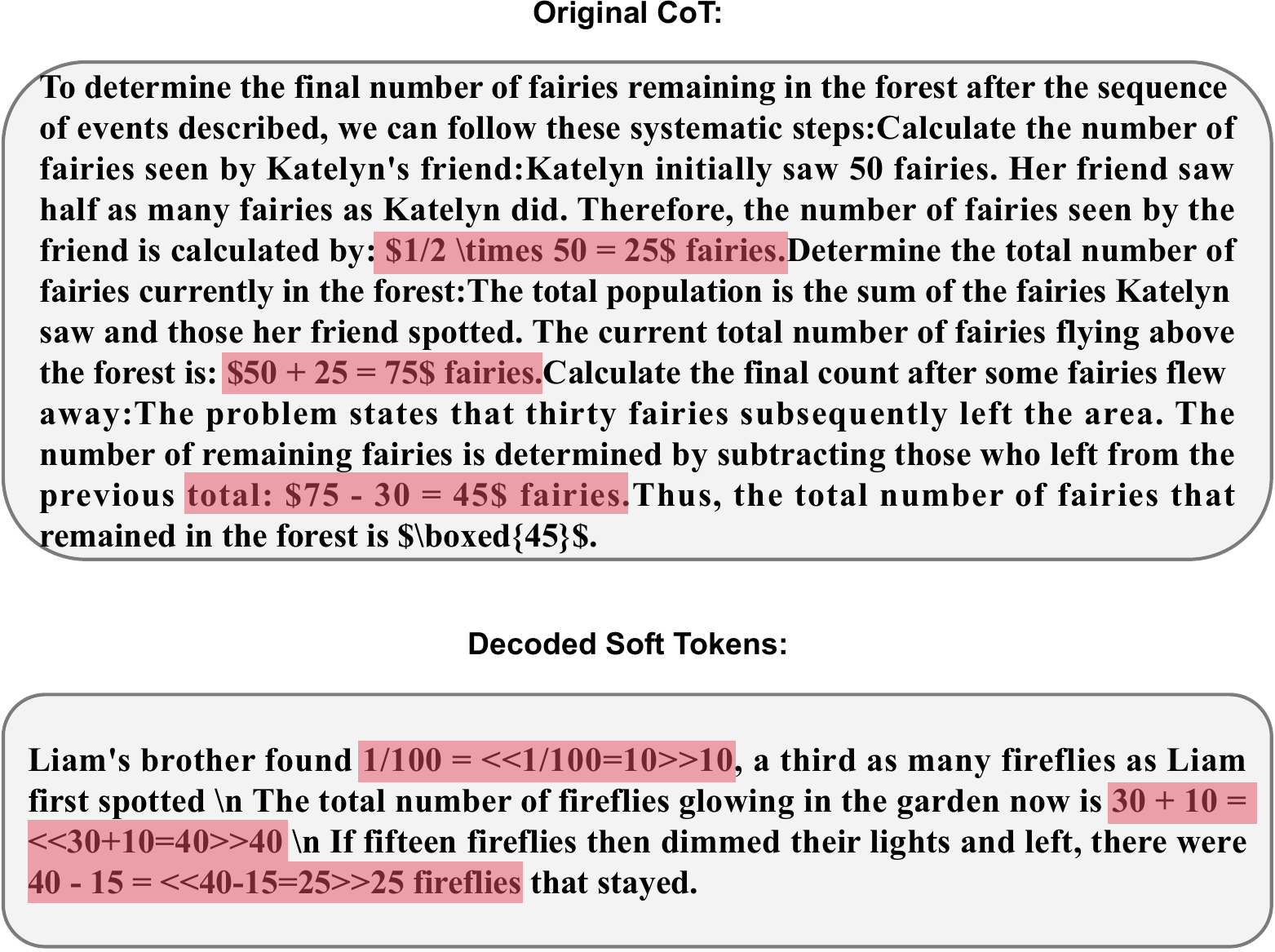} 
    }

    \caption{A case study on Math dataset. Output using Qwen2.5-7B-Instruct as the executor and Qwen2.5-1.5B as the compressor. The structural information about reasoning that appear simultaneously in both the original CoT and the decoded soft tokens are highlighted in red \raisebox{0.5ex}{\colorbox{MyHighlight}{\rule{-1pt}{0pt}}}.}
    \label{fig:case_ut_2} 
\end{figure*}

We present a case study of soft tokens with Qwen2.5-7B-Instruct as executors on the GSM8K and Math in Figure \ref{fig:case_ut_1} and Figure \ref{fig:case_ut_2}.

\subsection{Case Studies for UCoT}\label{C.8}
\begin{figure*}[h]
    \centering
    \resizebox{0.8\textwidth}{!}{%
        \includegraphics{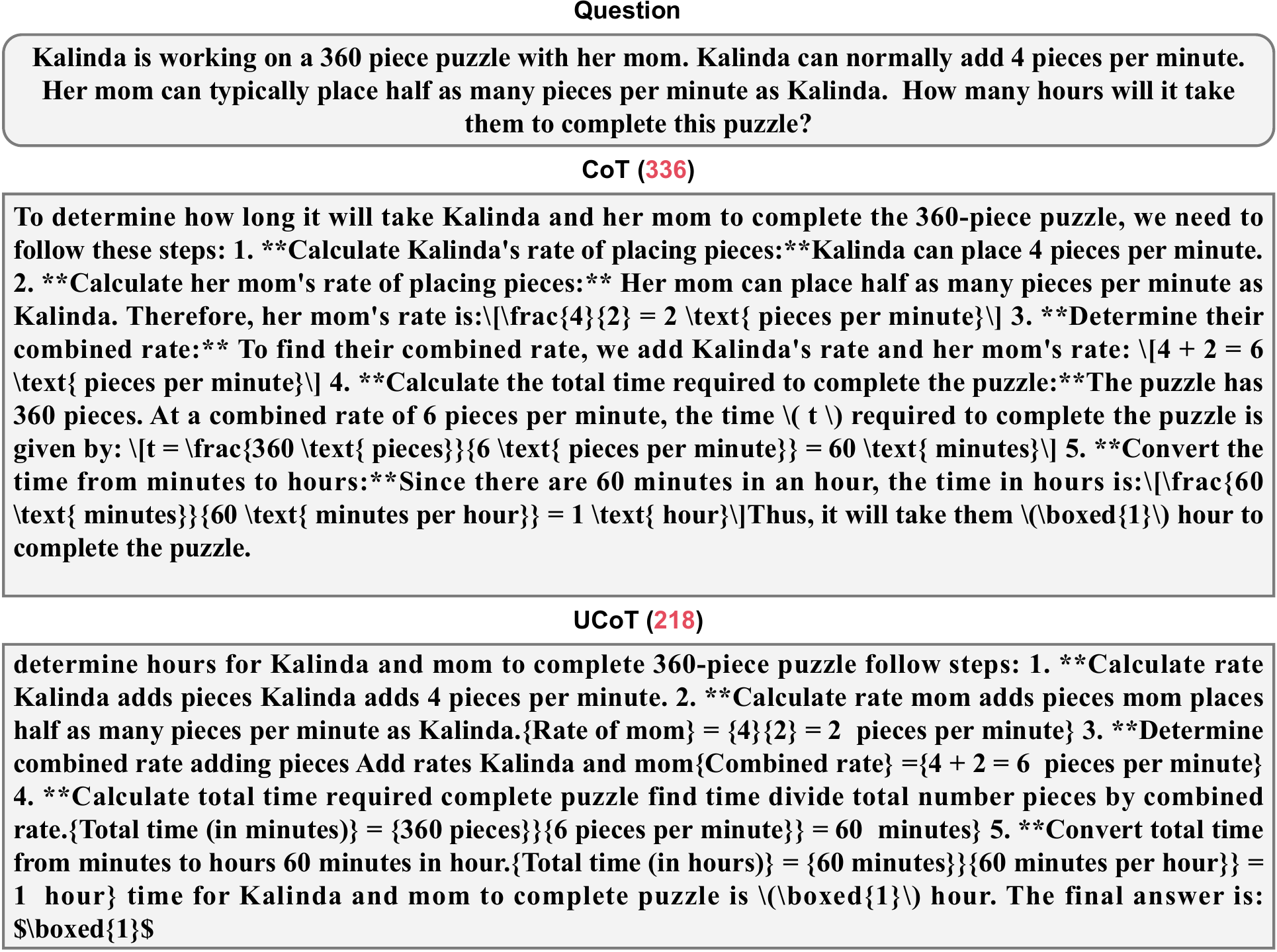} 
    }

    \caption{A case study on GSM8K dataset. Output using Qwen2.5-7B-Instruct as the executor with the CoT compression ratio of 0.7. We highlight the number of CoT tokens in red within parentheses.}
    \label{fig:case1} 
\end{figure*}

\begin{figure*}[!h]
    \centering
    \resizebox{0.8\textwidth}{!}{%
        \includegraphics{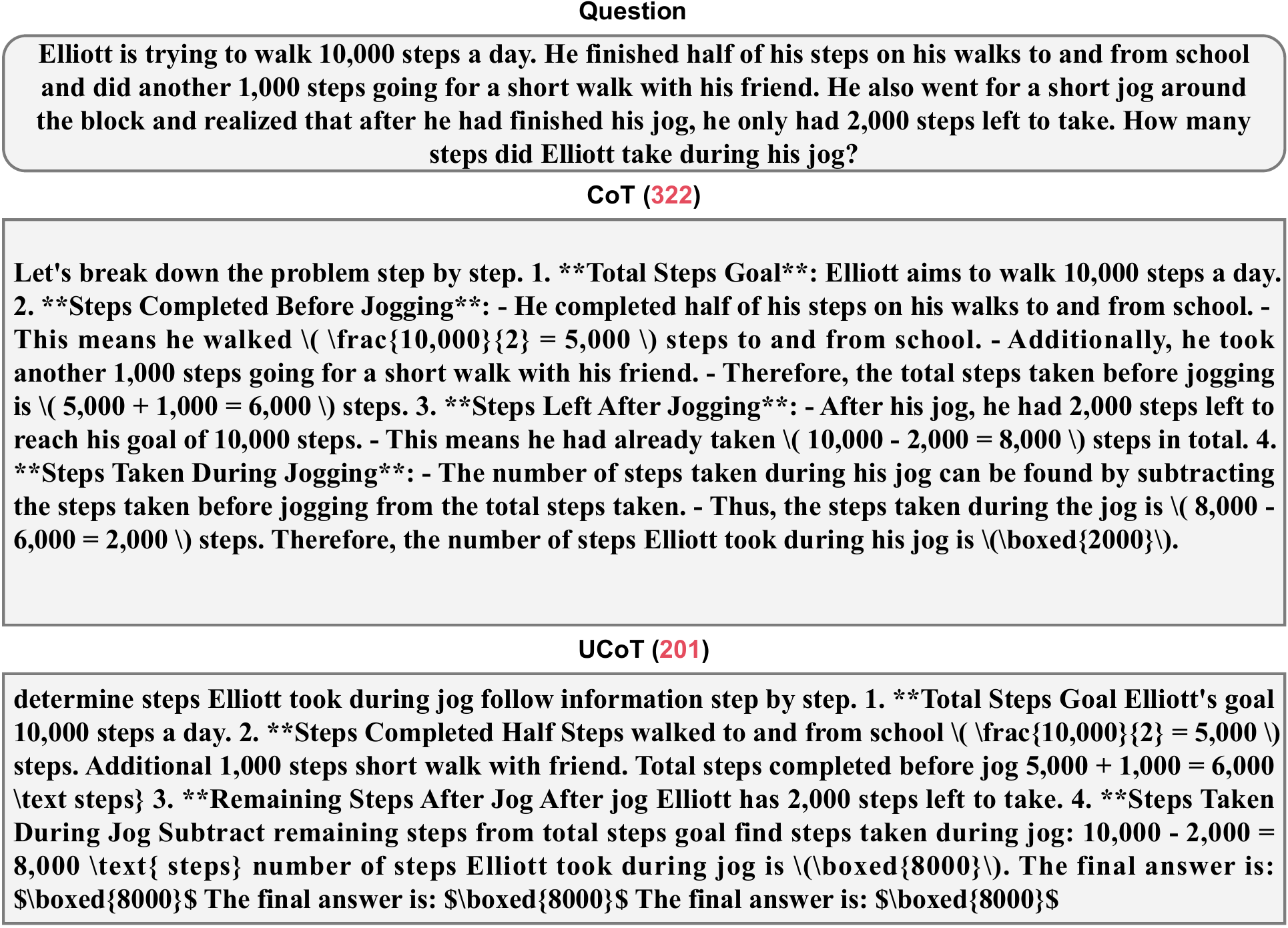} 
    }

    \caption{A case study on GSM8K dataset. Output using Llama-3.1-8B-Instruct as the executor with the CoT compression ratio of 0.7. We highlight the number of CoT tokens in red within parentheses.}
    \label{fig:case2} 
\end{figure*}

\begin{figure*}[tbp]
    \centering
    \resizebox{0.8\textwidth}{!}{%
        \includegraphics{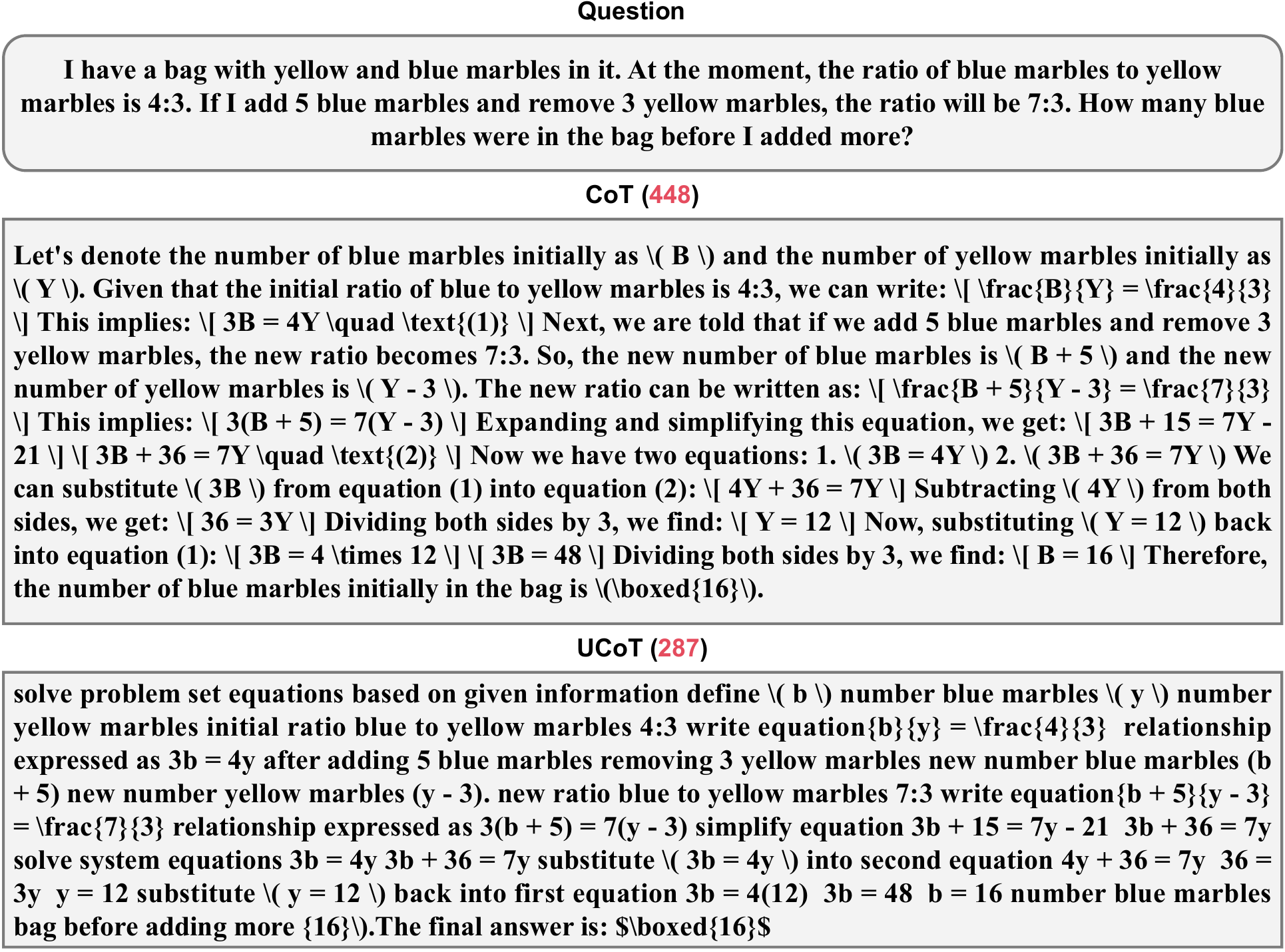} 
    }

    \caption{A case study on Math dataset. Output using Qwen2.5-7B-Instruct as the executor with the CoT compression ratio of 0.7. We highlight the number of CoT tokens in red within parentheses.}
    \label{fig:case3} 
\end{figure*}

\begin{figure*}[!h]
    \centering
    \resizebox{0.8\textwidth}{!}{%
        \includegraphics{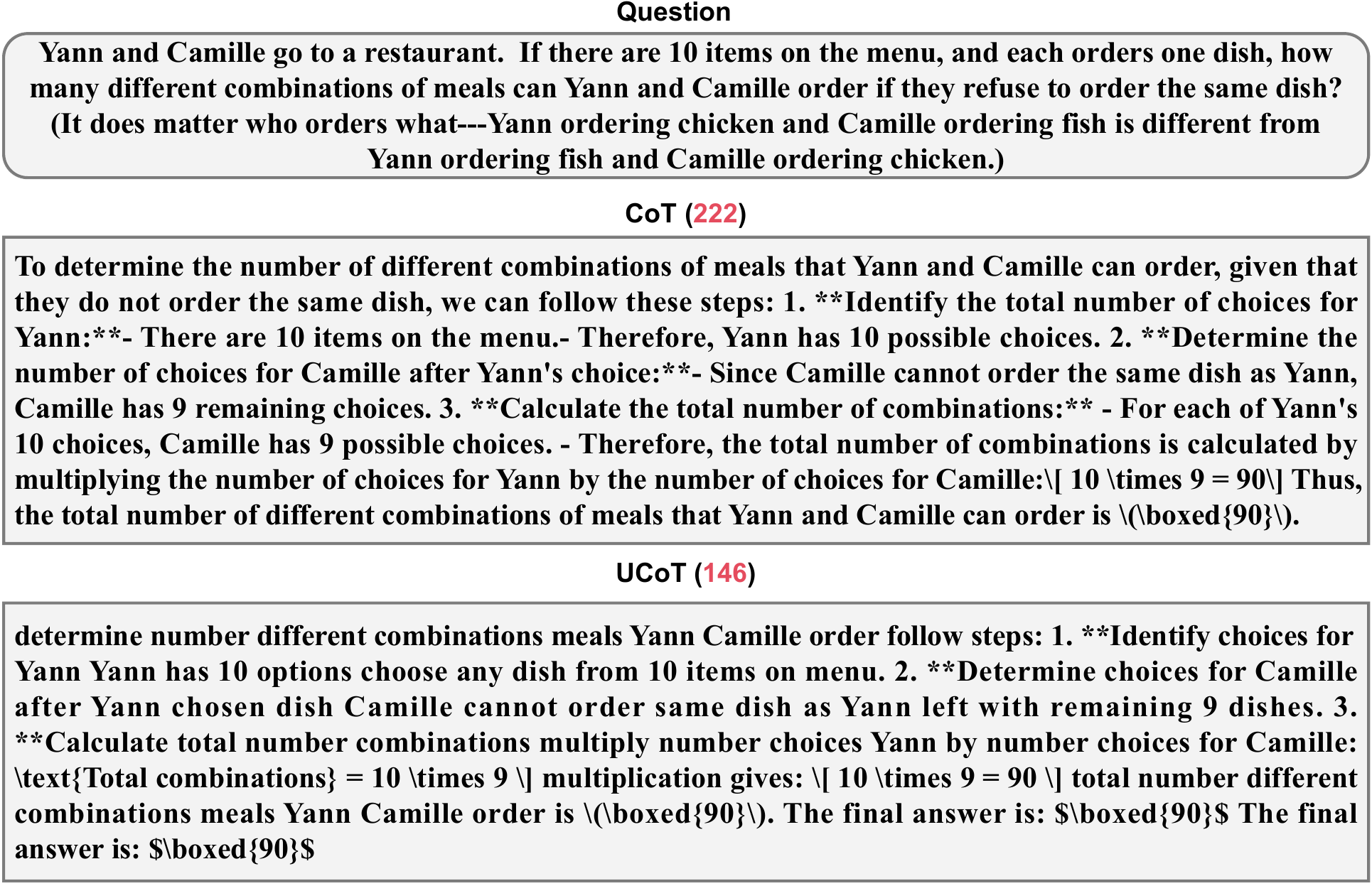} 
    }

    \caption{A case study on Math dataset. Output using Llama-3.1-8B-Instruct as the executor with the CoT compression ratio of 0.7. We highlight the number of CoT tokens in red within parentheses.}
    \label{fig:case4} 
\end{figure*}

In this section, we present a case study of UCoT with Qwen2.5-7B-Instruct and Llama-3.1-8B-Instruct as executors on the GSM8K and Math dataset. The experimental results are shown in Figures \ref{fig:case1} to \ref{fig:case4} where UCoT preserves the logical semantics in the original CoT prompts while significantly reducing the output length of the executor.

\vspace{-2.3cm}

\end{document}